\documentclass[journal]{IEEEtran}

\usepackage{times}
\usepackage{epsfig}
\usepackage{graphicx}
\usepackage{amsmath}
\usepackage{amssymb}
\usepackage{gensymb}
\usepackage{subcaption}
\usepackage{url}

\usepackage{array}
\usepackage{multirow}
\usepackage{textcomp}
\usepackage{array}
\usepackage{colortbl}
\graphicspath{{./graphics/}}
\usepackage[table]{xcolor}
\usepackage{booktabs}
\usepackage{multirow}

\usepackage{todonotes}
\usepackage[normalem]{ulem}
\usepackage{gensymb}

\newcommand{\eg}{{\it e.g.},~}
\newcommand{\ie}{{\it i.e.},~}
\newcommand{\etal}{{\it et al.}~}

\definecolor{myblue}{rgb}{0,0,1}
\definecolor{myred}{rgb}{0.8, 0, 0}
\definecolor{mygreen}{rgb}{0, 0.6, 0}

\colorlet{tableheadcolor}{blue!25} 
\colorlet{tablerowcolor}{blue!10} 
\begin{document}

\title{Iris Recognition After Death}

\author{Mateusz~Trokielewicz,~\IEEEmembership{Student Member,~IEEE,}
        Adam~Czajka,~\IEEEmembership{Senior Member,~IEEE,}
        and~Piotr~Maciejewicz
\thanks{{\bf Mateusz Trokielewicz} is with the Research and Academic Computer Network NASK, Warsaw, Poland.}
\thanks{{\bf Adam Czajka} is with the University of Notre Dame, IN, USA.}
\thanks{{\bf Piotr Maciejewicz} is with the Department of Ophthalmology, Medical University of Warsaw, Warsaw, Poland.}
\thanks{This paper has supplementary downloadable material available at http://ieeexplore.ieee.org, provided by the authors. The material includes an ISO/IEC-conformant analysis of post-mortem iris sample quality. Contact mateusz.trokielewicz@nask.pl for further questions about this work.}
}

\markboth{Manuscript accepted for publication in the IEEE TRANSACTIONS ON INFORMATION FORENSICS AND SECURITY.}%
{Trokielewicz \MakeLowercase{\textit{et al.}}: Iris Recognition After Death}


\maketitle

\begin{abstract}
This paper presents a comprehensive study of post-mortem human iris recognition carried out for 1,200 near-infrared and 1,787 visible-light samples collected from 37 deceased individuals kept in the mortuary conditions. We used four independent iris recognition methods (three commercial and one academic) to analyze genuine and impostor comparison scores and check the dynamics of iris quality decay over a period of up to 814 hours after death. This study shows that post-mortem iris recognition may be close-to-perfect approximately 5 to 7 hours after death and occasionally is still viable even 21 days after death. These conclusions contradict the statements found in past literature that the iris is unusable as biometrics shortly after death, and show that the dynamics of post-mortem changes to the iris that are important for biometric identification are more moderate than previously hypothesized. The paper contains a thorough medical commentary that helps to understand which post-mortem metamorphoses of the eye may impact the performance of automatic iris recognition. An important finding is that false-match probability is higher when live iris images are compared with post-mortem samples than when only live samples are used in comparisons. This paper conforms to reproducible research and the database used in this study is made publicly available to facilitate research on post-mortem iris recognition. To our knowledge, this paper offers the most comprehensive evaluation of post-mortem iris recognition and the largest database of post-mortem iris images.
\end{abstract}


\section{Introduction}
\IEEEPARstart{I}{dentification} of deceased individuals through their biometric traits has long been used for forensic purposes, exploiting characteristics such as fingerprints, DNA, or dental records to recognize victims of accidents, or natural disasters and crimes \cite{WaymanJainBiometricSystemsBook2005, JainRossBiometricsForensics2015}. Post-mortem iris recognition, however, has not received considerable attention, despite excellent performance of this method when applied to live eyes. Studying this unfamiliar area has at least two important goals:

\paragraph{To aid forensics} Can iris biometrics be a fast and accurate complement or alternative method to the existing approaches to post-mortem identification? If the answer is affirmative, it could be useful in cases when other methods cannot be applied, such as for victims of accidents with severed fingers or disfigured faces.

\paragraph{To improve security} Can dead iris be effectively used in presentation attack? Understanding the dynamics and reasons for post-mortem iris performance degradation allows to provide more precise answer to this question, and may help in development of countermeasures against forgeries with cadaver eyes.

To come up with as many answers as possible, this paper presents a comprehensive feasibility study of post-mortem iris recognition involving iris images acquired from 5 hours to almost 34 days after death in near-infrared (NIR) and visible-light (VIS). It is centered around the following six questions:

\begin{enumerate}
	\item \emph{Is automatic iris recognition possible after death?}
	\item \emph{What are the dynamics of deterioration in iris recognition performance?}
	\item \emph{What type of images are the most favorable for post-mortem iris recognition?}
	\item \emph{What are the main reasons for errors when comparing post-mortem iris samples?}
	\item \emph{Which factors influence post-mortem iris recognition performance?}
	\item {\it What are the false-match risks when post-mortem samples are compared against databases of live iris images?}
\end{enumerate}

To answer the above questions we acquired 1,200 NIR and 1,787 VIS images from 37 cadavers during multiple sessions organized from 5 to 814 hours after death. The bodies were kept in controlled mortuary conditions and stable temperature of 6\degree~Celsius (42.8\degree~Fahrenheit). Four independent iris recognition methods were used to show that automatic iris recognition stays occasionally viable even 21 days after death, and is close to perfect approximately 5 to 7 hours post-mortem. This allows to reject prior hypotheses that the iris cannot be used as biometrics after death \cite{DaugmanPostMortem, SaeedPostMortem, IrisGuardPostMortem, IriTechPostMortem}. In this paper we also show that using the red channel of VIS post-mortem iris images can be considered as a good alternative to NIR samples. Images consisting of only red channel will be later referred to as `R images' in the paper. We also show that the performance of cross-wavelength post-mortem iris matching (NIR vs R) is significantly worse than same-wavelength (NIR vs NIR and R vs R) matching. We analyze possible reasons for false match and false non-match instances, and by manual correction of the segmentation for the whole dataset we assess the impact of erroneous segmentation on the post-mortem iris recognition performance. The paper provides medical commentary on these post-mortem metamorphoses observed in the eye that degrade the recognition reliability the most. We discuss briefly relation of gender, age, and cause of death with post-mortem iris recognition. To our knowledge, this paper comprises the most extensive and comprehensive study regarding post-mortem iris recognition, and offers the largest dataset of iris images collected from deceased subjects.

The paper is organized as follows. Section \ref{sec:Related} provides an overview of publications and claims related to post-mortem iris recognition. Section \ref{sec:Database} describes the database used in this study, details of the collection protocol, its timeframe, the data statistics, and gives information on obtaining a copy of the dataset. Section \ref{sec:Medical} contains a medical commentary on the expected post-mortem changes to the eye, according to current academic knowledge. Experiments involving four different, commercial and academic iris recognition methods are reported on in Section \ref{sec:Experiments}. Conclusions answering six questions posed above and discussing limitations of this study are provided in Section \ref{sec:Conclusions}.

\section{Related Work}
\label{sec:Related}

A belief that iris recognition is difficult or even not feasible after a person's death has been hypothesized for a long time in both scientific and industry communities. In 2001, John Daugman, who without doubt can be referred to as 'the father of iris recognition', stated the following in his interview for the BBC: \emph{'soon after death, the pupil dilates considerably, and the cornea becomes cloudy'}. While this statement is fairly moderate, others put forward far stronger claims regarding post-mortem iris biometrics, for instance, Szczepanski \etal write that \emph{'the iris (...) decays only a few minutes after death'} \cite{SaeedPostMortem}. References to post-mortem iris recognition can be also found in commercial materials, for instance: \emph{(...) the notion of stealing someone's iris after death is \textbf{scientifically impossible}. The iris is a muscle; it completely relaxes after death and results in a fully dilated pupil with no visible iris at all. \textbf{A dead person simply does not have a usable iris!}'} \cite{IrisGuardPostMortem}, or \emph{'after death, a person's iris features will vanish along with pupil's dilation'} \cite{IriTechPostMortem}. However, none of these assertions are backed by any scientific argumentation or experimentation.

Due to technical and ethical difficulties in collecting biometric samples from cadavers, only a small number of researchers have studied the post-mortem iris recognition problem using scientific methods. Sansola \cite{BostonPostMortem} used IriShield M2120U iris recognition camera together with IriCore matching software in her experiments involving 43 subjects who had their irises photographed at different post-mortem time intervals. Depending on the post-mortem interval, the method yielded 19-30\% of false non-matches and no false matches. She reported a relationship between eye color and post-mortem comparison scores, with blue/gray eyes yielding lower correct match rates (59\%) than brown (82\%) or green/hazel eyes (88\%). Saripalle \etal \cite{PostMortemPigs} used ex-vivo eyes of domestic pigs and they came to the conclusion that irises are slowly degrading after being taken out of the body, and lose their biometric capabilities 6 to 8 hours after death. However, ex-vivo eye degradation is expected to be much faster than the same processes occurring while the eye is still a part of the cadaver. Ross \cite{RossPostMortem} observed a fadeout of the pupillary and limbic boundaries found in post-mortem iris images, as well as corneal opacity, which developed in all of the samples under observation.

In our previous work we showed that despite popular claims, the iris can still successfully serve as a biometric identifier for 27 hours after death \cite{TrokielewiczPostMortemICB2016}. The pupils were found to remain in the so called 'cadaveric position', meaning that no excessive dilation or constriction is present, and hence the iris structure remains well visible. About 90\% of the irises were correctly recognized when photographed a few hours after death. As time after death increases, the equal error rate drops to 13.3\% when images captured approximately 27 hours after death are compared against those obtained 5h after demise. Later, we showed that correct matches can still be expected even after 17 days \cite{TrokielewiczPostMortemBTAS2016} and offered the first known to us database of 1330 NIR and VIS post-mortem iris images acquired from 17 cadavers \cite{WarsawColdIris1}.

Bolme \etal \cite{BolmePostMortemBTAS2016} attempted to track biometric capabilities of face, fingerprint and iris during human decomposition. Twelve subjects were placed in the outdoor conditions to assess how the environment and time affect the biometric performance. Although fingerprints and face are shown to be moderately resilient to decomposition, the irises degraded quickly regardless of the temperature. The authors state that irises typically became useless from the recognition viewpoint only a few days after exposition to outdoor conditions, and if the bodies are kept outside for 14 days the correct verification decreases to only 0.6\%. The real-life chance of recognizing an iris is estimated by the authors to be even less than 0.1\%. The most recent paper in this field by Sauerwein \etal \cite{Sauerwein_JFO_2017} showed that irises stay readable for up to 34 days after death, when cadavers were kept in outdoor conditions during winter. The readability was assessed by human experts acquiring the samples and no iris recognition algorithms were used in this study, however it suggests that winter conditions increase the chances to see an iris even in a cadaver left outside for a longer time.

\section{Database of Post-Mortem Iris Images}
\label{sec:Database}

\subsection{Data Collection}

A crucial part of this study was to create a new database of iris images, which would represent eye regions of recently deceased persons. We had a rare opportunity to collect iris scans from hospital mortuary subjects. The following section briefly characterizes the acquisition methodology and timeline of acquisition sessions.

\subsubsection{Equipment} Two different sensors were used for image acquisition: a commercial iris sensor operating in NIR light \textbf{IriShield M2120U}, and a consumer-grade color camera \textbf{Olympus TG-3}. Color images were collected simultaneously with NIR ones and each subject and each acquisition session are represented by at least one image of each type. The IriShield sensor is equipped with a near-infrared illuminant, whose irradiance falls into the 710-870 nm band, with a peak at 810 nm \cite{IriCoreIlluminationWavelength}.

\subsubsection{Environmental Conditions} All acquisition sessions were conducted in the hospital mortuary. The temperature in the mortuary room was approximately 6\degree~Celsius (42.8\degree~Fahrenheit). Other conditions, such as air pressure and humidity were unknown, yet stable. The environmental conditions, in which the cadavers were kept prior to entering the cold storage are unknown. 
 
\subsubsection{Acquisition Timeframe} 
\begin{figure}[!htb]
	\centering
	\includegraphics[width=0.49\textwidth]{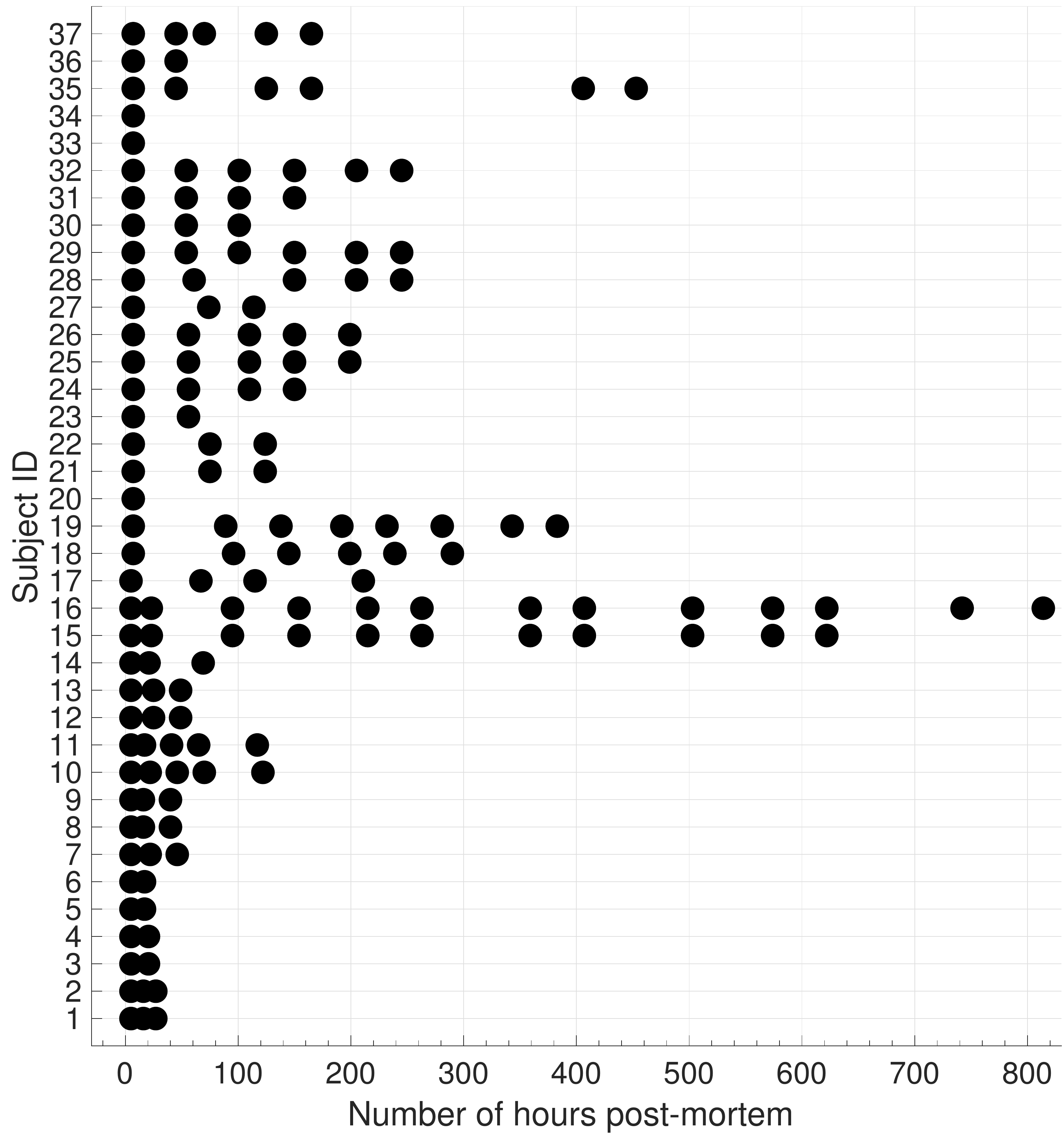}
	\caption{Hours post-mortem for each acquisition session plotted independently for each deceased subject.}
	\label{fig:sessions}
\end{figure}

From 1 to 13 acquisition sessions could be organized for a given subject in this study. In each session at least one NIR and one VIS image were acquired. Due to ethical concerns, no ante-mortem samples could be collected, hence the first session for each subject was always organized as soon after death as possible, typically 5 to 7 hours. The following sessions were organized based on the availability of deceased persons, who were subject to medical or police investigations, and were retained in the mortuary during varying time slots. The overview of acquisition sessions for all subjects is shown in Fig. \ref{fig:sessions}. For three subjects, namely 20, 33, and 34, only a single acquisition session was possible.

\subsubsection{Within-Session Acquisition Protocol} When collecting images within a single acquisition session, all samples can be considered separate presentations as recommended by the ISO/IEC 19795-2, \ie after taking a photograph, the camera was moved away from the subject and then positioned for the next acquisition.

\subsection{Quality of the Samples}
For each image collected for this database, we have implemented the calculation of two quality metrics suggested in the ISO/IEC standard on iris image quality \cite{ISO2}, namely: \emph{grayscale\_utilisation} and {\it sharpness}, as these are the only metrics from the current standard that do not require ground truth segmentation information, but rather require a raw iris image. Formulae for these calculations, statistical results, and quality scores for selected images are included in the Supplementary Materials.

\subsection{Statistics}
\label{sec:DatabaseStats}
The dataset is a significant extension over the corpus used in our previous studies, comprising 1,200 NIR images, accompanied by 1,787 color images. These images represent eye regions of 37 different subjects (73 different irises, since only one eye was imaged for one cadaver). Age of the deceased ranged from 19 to 75 years old. 5 subjects were female and 32 were male. Causes of death included heart failure (18 subjects), car crash (7), suicide by hanging (7), murder (1), poisoning (2), and head trauma (2). The eye colors were blue/gray/light green (29 cadavers), light brown/hazel (5) and dark brown (3). Detailed description for each subject can be found in the metadata accompanying the released database.

\subsection{Database Access}
To conform with the reproducibility guidelines and facilitate research in this area, the database used in this study is made available along with the paper. A copy can be requested at: \linebreak \url{http://zbum.ia.pw.edu.pl/EN}~$\rightarrow$~\url{Research}~$\rightarrow$~\url{Databases}~$\rightarrow$ \linebreak {\bf Warsaw-BioBase-Post-Mortem-Iris v2.0}.

\section{Medical Background}
\label{sec:Medical}
\subsection{Post-Mortem Changes to The Eye}

\subsubsection{General overview} Initially, the post-mortem decomposition of human organs may not be visible to the naked eye, since these processes start at the cellular level and then slowly progress to the macroscopic level. Early changes include algor mortis (body cooling), rigor mortis (desiccation with stiffening of the body), pallor mortis (paleness) and livor mortis (lividity), while late ones comprise of progressing decomposition caused by autolysis and putrefaction. Autolysis is a cellular self-destruction process caused by hydrolytic enzymes that were originally contained within cells. Putrefaction is a degradation of tissue caused by microorganism (\eg bacterial) activity, and is visible macroscopically as discoloration or bloating of the skin.

\subsubsection{The cornea} The most prominent metamorphoses observed in the eyes after death, and possibly the most troubling for iris recognition, are the changes to the cornea. A live cornea is a clear, transparent, dome-shaped structure in front of the eyeball. It is responsible for about 2/3 of the total eye optical power because of its curvature and the resulting refractive index. The cornea must remain transparent to refract light properly, but also to allow good quality iris image capturing. Its transparency is maintained by a controlled hydration with the tear film, produced by lacrimal glands and distributed by eyelids. As secretion stops, anoxia, dehydration and acidosis lead to progressing autolysis of the cells. Corneal thickness decreases immediately after death and increases thereafter. This results in opacification that increases with time. Upon death the cornea slowly becomes hazy. The change in corneal opacity is believed to be secondary to the change in hydration and architectural destruction of the collagen fiber network, functional alteration of corneal endothelium, disregulation of proteoglycan hydration and ion concentration in corneal stroma. It was confirmed that temperature has significant influence on protein degradation. Another effect associated with these mechanisms is the wrinkling of the corneal surface, manifesting itself with difficulties to obtain a good visibility of the underlying iris pattern. The progression of these effects is influenced by multiple factors, such as closure of the eyelids, environment humidity, temperature, and air movement. It is also dependent on the age and general medical condition of a deceased person. Due to reduced intraocular pressure, we can also notice central depression of the globe, flaccidity of the eyeball, and loss in its firmness \cite{Prieto2015}.

\subsubsection{The iris} There are no evident changes to the iris surface observed after death. After demise, pupils are usually mid-dilated (a.k.a. `cadaveric position'), and in some cases they can be slightly dilated, because of the relaxation of the iris muscles and later they can become slightly constricted with the onset of rigor mortis of the constrictor muscles. In other cases, we may observe initial myosis within the first few hours after death with strong variations between individual cases. If rigor mortis affects ciliary muscles of two irises unequally, pupils in both eyes may have different apertures. Sometimes, if different segments of the same iris are unequally affected then the pupil may be irregularly oval or have an eccentric position. Shape and size of the pupils can also depend on the medical history of the subject, including treatment with drugs and eye surgeries. 

\subsubsection{Muscles of the iris} Death was once defined as the cessation of heartbeat and breathing, but with the development of cardiopulmonary resuscitation these can be restarted in some cases. Thus, we now typically rely upon the concept of brain death to define whether a person is clinically dead. Supravitality -- sensitivity to excitation -- relates to survival rates of tissue after complete irreversible ischemia (restriction of blood flow). The supravital reaction is the response of muscles to stimulation in the early period after death, as some cells do not die immediately after the brain death. Within the first couple of hours we can notice a decreasing pupillary reaction for pupillomotoric drugs, for example to pilocarpine and atropine \cite{LARPKRAJANG20161}. Iris tissue response to myotic (pupil-constricting) and mydriatics (pupil-dilating) agents can be observed. 

\subsubsection{Other aspects} There are some other changes that the eye undergoes after death. The loss of intraocular lens transparency with time is due to the metabolic processes that take place between the lens and the aqueous and vitreous humor, and the aggregation of crystalline proteins in the fibers of the lens nucleus. It has been hypothesized that lens proteins aggregate to large particles that scatter light, causing lens opacity \cite{Prieto2015}. After death we may observe a black spot in the sclera, referred to as `tache noire', caused by desiccation of the sclera with open eyelids, usually symmetrical corresponding to the position of the eyelids. Also, the vitreous humor -- gelatinous substance contained in the posterior chamber of the eye, keeping the retina in place and maintaining the spherical shape of the eyeball -- tends to liquefy, and later to dry, starting the process of eyeball collapse \cite{Belsey2016}.

\subsection{Visual Inspection of Post-Mortem Changes} 
\label{sec:VisualInspection}

We have taken the effort to carefully examine the samples throughout the time period since death for all subjects, and confront the observed changes with medical knowledge. This yielded a qualitative evaluation of post-mortem changes to the iris reported in this Section. Having both NIR and VIS images is crucial for such assessment, as these two types of illumination often reveal different appearance of the iris when changes to the cornea and the anterior chamber are present. This is shown in Fig. 3, where visible-light samples are compared against near-infrared samples for the same eye. Such differences are also reported on in the works of Aslam \etal \cite{Aslam} and Trokielewicz \etal \cite{TrokielewiczDiseasesIMAVIS} related to the disease influence on iris recognition performance. Both studies show that the NIR illumination typically used in iris recognition cameras is capable of alleviating corneal opacification effects to some extent.

\begin{figure*}[!htb]
	\centering
	\includegraphics[width=\textwidth]{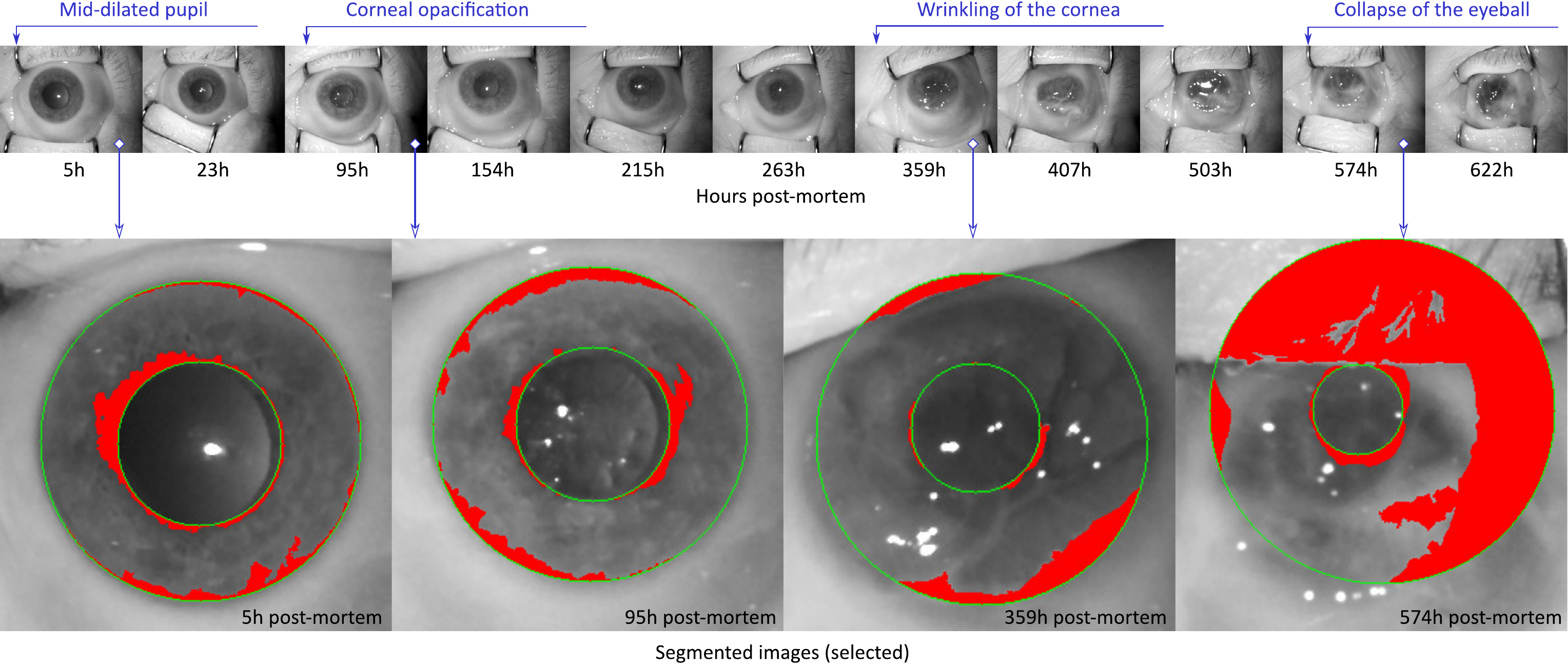}
	\caption{Example measurements for a single subject throughout the period of time post-mortem. Symptoms that are expected to appear as time since death elapses are denoted above the top row. OSIRIS segmentation results are also shown for selected images, presenting the degradation of iris image segmentation quality with passing time. Red-colored portions of the image are denoted by the algorithm as not representing the iris and therefore do not participate in comparison.}
	\label{fig:coldSamples}
\end{figure*}

A summary of example post-mortem changes that appear in the eye is presented in Fig. \ref{fig:coldSamples}, together with a timeframe for a selected subject. It must be noted, however, that the dynamics of these changes are heavily subject-dependent and can happen with different rapidity, intensity and prevalence on the appearance of iris tissue.

First, a corneal opacification progresses with time since death, and it becomes visible after a few days post-mortem (\eg 95 hours, or 4 days, after death, as depicted in Fig. \ref{fig:coldSamples}). Second, a wrinkling of the corneal surface is expected to appear (\eg 359 hours, or 15 days, as shown in Fig. \ref{fig:coldSamples}). At this point, a strong influence on the automatic image segmentation procedures can be anticipated, as the iris tissue becomes less visible and additional patterns and light reflections emerge. Third, a loss of intraocular pressure in the eyeball due to post-mortem biochemical changes can be observed (\eg 574 hours, or 24 days, as illustrated Fig. \ref{fig:coldSamples}), causing the eye to slowly collapse into the eye socket. At this point in time, iris recognition methods are expected to seldom work, as the iris pattern is severely obstructed and thus challenging for iris image segmentation. Finally, after about a month, the eyeball was observed to dry out completely, leaving no traces of a healthy iris structure.

\begin{figure}[!tb]
	\centering
	\includegraphics[width=0.48\textwidth]{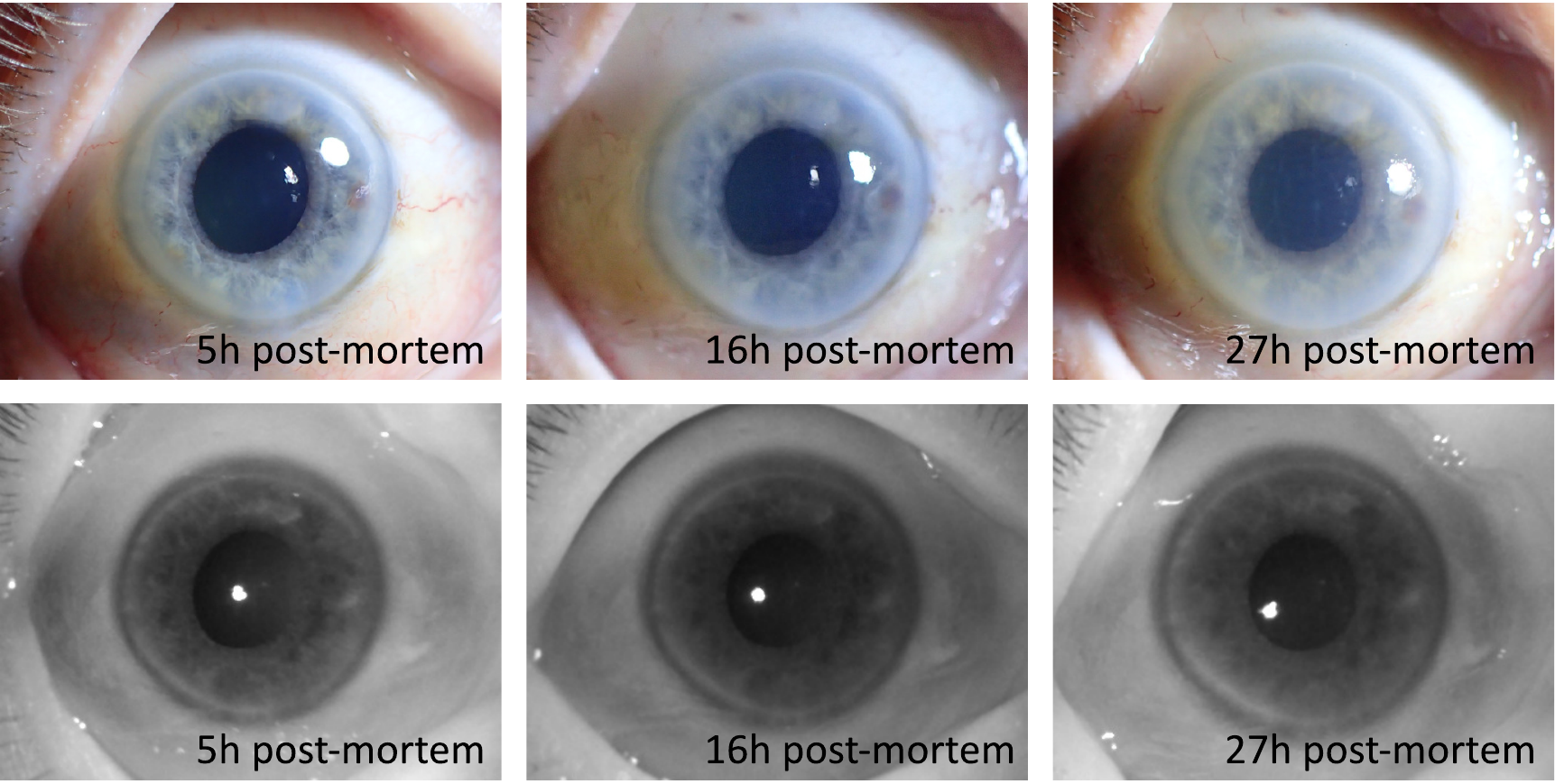}\vskip1mm
	\includegraphics[width=0.48\textwidth]{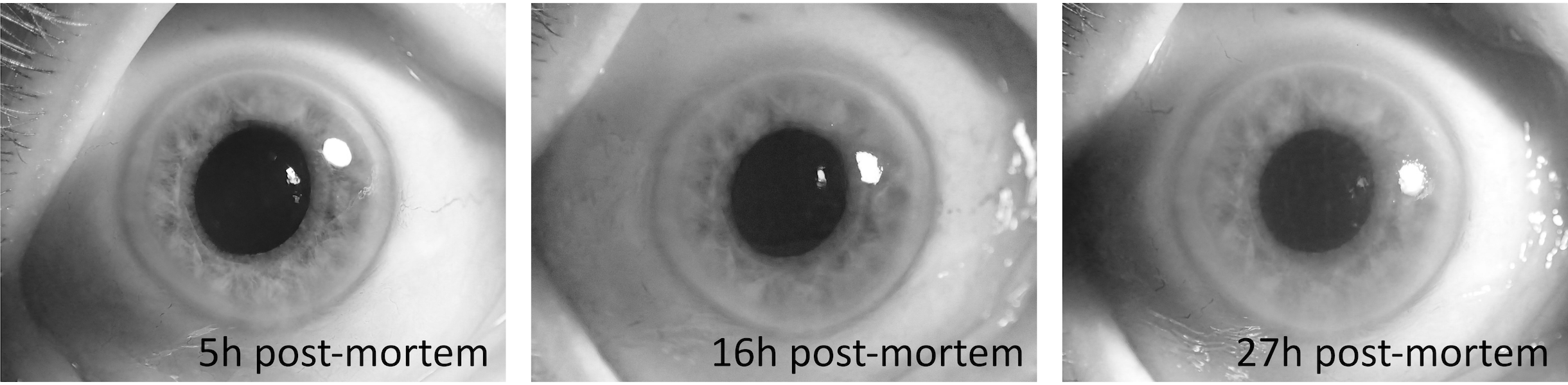}
	\caption{From left to right: \textbf{visible-light (VIS)} (top row),  \textbf{near-infrared (NIR)} (middle row), and \textbf{red channel (R)} (bottom row) images obtained in the first three sessions: \textbf{5}, \textbf{16}, and \textbf{27 hours} after death. Progressing corneal opacity can be spotted in visible-light images, but NIR imaging seems to be more insensitive to corneal haze.}
	\label{fig:NIR-VIS-samples}
\end{figure}

Contrary to initial predictions, we did not come across any sample that would be affected by \emph{tache noire}. Also, the severe corneal opacification was visible in original VIS samples only, while NIR and R images worked in favor of exposing post-mortem iris texture better than original VIS samples, as depicted in Fig. \ref{fig:NIR-VIS-samples}.

\section{Experimental Study}
\label{sec:Experiments}

\subsection{Iris Recognition Methods}
For a comprehensive analysis of how iris recognition can perform when used with post-mortem samples, we have employed four independent iris recognition methods. Three of them are commercially available products, and one is an open source solution.

\subsubsection{VeriEye} This commercial product is offered by Neurotechnology in the form of the Software Development Kit (SDK) \cite{VeriEye}. The manufacturer does not divulge  algorithm details, apart from the claim that off-axis iris localization is employed with the use of active shape modeling. The algorithm has been evaluated by NIST in their ICE 2005 \cite{ICE2005} and IREX \cite{IREXgeneral} projects. VeriEye is the only algorithm in this study which returns a similarity score between two iris images, rather than a difference score. Hence, the higher the score, the better the match between samples. Scores between 0 and 40 denote different-eye (impostor) pairs, while scores above 40 are expected for same-eye (genuine) pairs.

\subsubsection{IriCore} Similarly to the VeriEye method, the IriCore matcher is offered commercially as the SDK \cite{IriCore}. IriTech Inc. does not disclose any details on the underlying algorithm. The software is claimed by the manufacturer to  conform with the ISO/IEC 19794-6 standard \cite{ISO} and has been shortlisted by NIST in 2005 \cite{ICE2005} as one of the best iris recognition solutions. IriCore returns dissimilarity score from 0.0 to 2.0, with same-eye (genuine) scores expected to fall in between 0 and 1.1, and different-eye (impostor) scores between 1.1 and 2.0.

\subsubsection{MIRLIN} (\emph{Monro Iris Recognition Library}) is a third method offered on the market in the form of an SDK by FotoNation Ltd (formerly Smart Sensors Ltd) \cite{MIRLIN}. The underlying algorithm employs discrete cosine transform (DCT) calculated for overlapping iris image patches to deliver binary iris features \cite{Monro2007}. Similarly to Daugman's original method, the resulting iris codes are compared using exclusive or (XOR) operation and normalized by a number of valid bits (corresponding to iris portions that are not occluded), yielding a fractional Hamming distance. Comparing two images of the same eye should result in a score close to zero, while the distance between images of two different irises is expected to oscillate around 0.5.

\subsubsection{OSIRIS} \emph{Open Source for IRIS}, an academic solution developed within the BioSecure EU project \cite{OSIRIS}, has been open-sourced by its authors. Its principles follow the original works of John Daugman, with iris image segmentation and subsequent normalization to dimensionless polar coordinate system. A binary iris code is calculated using phase quantization of the Gabor filtering outcomes. Similarly to the MIRLIN method, the fractional Hamming distance between the codes is used as a comparison score. Values close to zero are expected for two same-eye images, while scores oscillating around 0.5 should be produced when two different-eye images are compared. However, due to compensation of the eyeball rotation, different-eye score distributions will more likely be skewed toward 0.4 -- 0.45 range. We introduced a modification of the original OSIRIS method to include score normalization as proposed by Daugman \cite{Daugman2007NewMethods}:

$$HD_{norm} = 0.5 - (0.5 - HD_{raw}) \sqrt{\frac{n}{N}}$$

\noindent
This transforms the samples of scores obtained when comparing different eyes into samples drawn from the same binomial distribution, as opposed to drawing sample scores from different binomial distributions with $\sigma$ dependent on the number of bits $n$ that were available for comparison (commonly unmasked bits). $N$ is the `typical number of bits compared (unmasked) between two different irises,' being said to equal 911 or 960, depending on the data. The $N$ parameter is estimated for a particular database of iris images. Since post-mortem iris samples are different from images of live iris images, and the OSIRIS does not necessarily use identical Gabor wavelets as used in [29], we estimated $N$ with our post-mortem samples, which equals to 1416 and 1446 for the automatic and manual segmentation, respectively. The total number of bits in the OSIRIS code is 1536.

\subsection{Database of Iris Images}

\subsubsection{Near-infrared and visible-light samples}
During data acquisition, we had the opportunity to collect images using two types of cameras: one producing near-infrared images of VGA size ($640\times480$ pixels), and the second producing color photographs of high resolution. Both near-infrared and visible-light images were used in visual inspection presented in Sec. \ref{sec:VisualInspection}. Using only red channel of color iris samples acquired in visible-light has been found to offer high recognition accuracy \cite{TrokielewiczBartuziJTIT} even when being matched with near-infrared samples \cite{TrokielewiczVisibleISBA2016}. Thus, in this study, two types of samples are used: a) original near-infrared and compliant to ISO/IEC 19794-6 and ISO/IEC 29794-6 standards, and b) red channel of visible-light images manually center-cropped to conform the VGA image type, as defined in ISO/IEC 19794-6. Hence, the resolution of all images is $640\times480$ pixels. This cropping of visible-light sample additionally protects the identity of donors, as original high-resolution visible-light images contained significant portions of face region. An example pair of near-infrared sample and the cropped red-channel image of the same eye is shown in Fig. \ref{fig:database-variants}.

\subsubsection{Manually-annotated ground-truth iris masks}
\begin{figure}[!htb]
	\centering
	\includegraphics[width=0.12\textwidth]{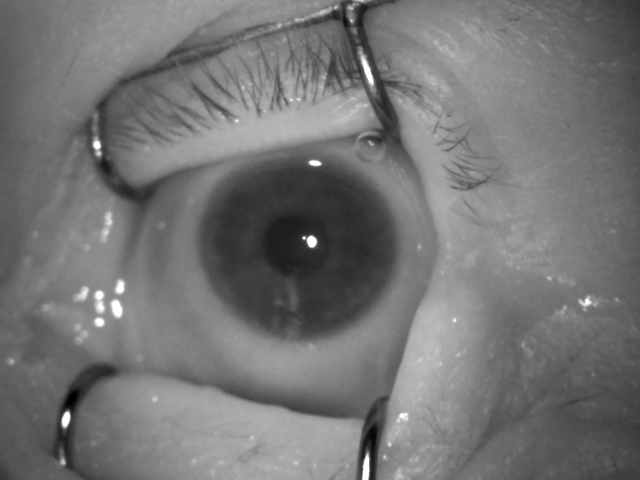}\hskip0.5mm
	\includegraphics[width=0.12\textwidth]{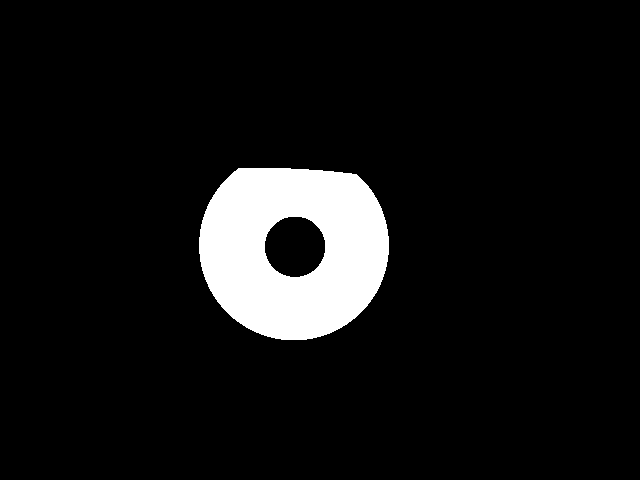}\hskip0.5mm
	\includegraphics[width=0.12\textwidth]{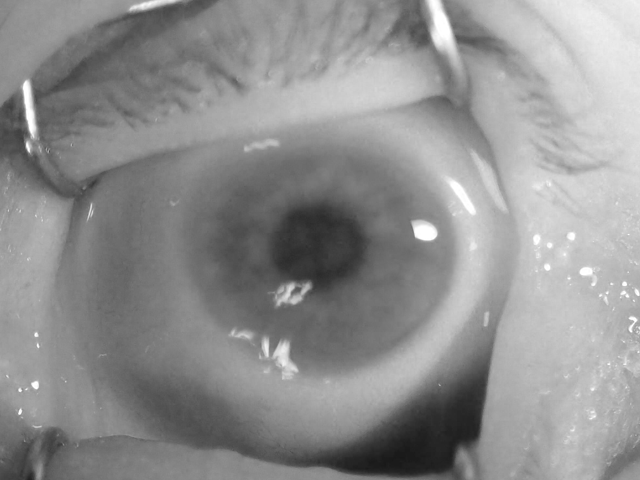}\hskip0.5mm
	\includegraphics[width=0.12\textwidth]{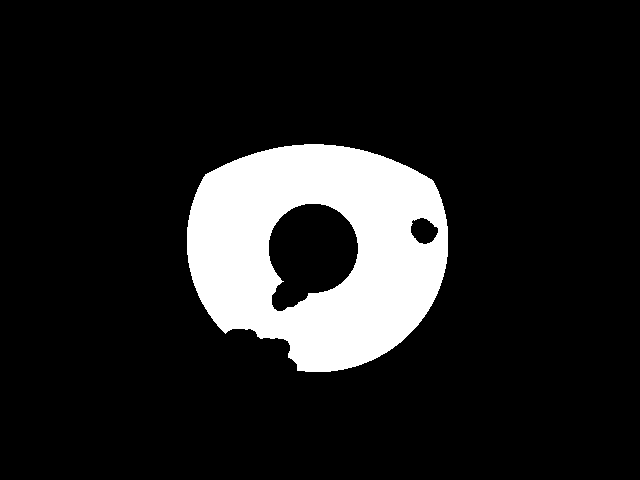}\hskip0.5mm
	\caption{Example NIR and R images of the same eye (65 hours after death) and the corresponding manually annotated masks.}
	\label{fig:database-variants}
\end{figure}

The erroneous execution of the segmentation stage is usually a main cause of drops in iris recognition performance, when samples presented to the algorithms are of challenging nature. To test whether this is also the case for post-mortem data, we have taken the effort to prepare manually annotated iris masks for all of the iris images involved in this study, including both NIR and visible light samples, as depicted in Fig. \ref{fig:database-variants}.

\subsubsection{Dataset of live iris images}
For the purpose of assessing false-match risks in scenarios when post-mortem samples are expected to be compared in an open-set scenario with the existing datasets of live iris images, we have collected a complementary dataset of iris images from living 74 subjects. To minimize the bias in the data, we have used the same sensor (IriShield MK 2120U) as used for the collection of post-mortem data. Two subject-disjoint subsets gathering data from 37 subjects each were created, containing 557 and 611 images, respectively.

\subsection{Types of Analyses}

Due to difficult data acquisition resulting in rather sparse and irregular image capture moments, the experiments conducted in this study are carried out three-fold: (i) a {\bf short-term analysis} that is based on samples collected in the first acquisition session for each subject, (ii) a {\bf long-term analysis} that employs the entire data representing a maximum period of 814 hours after death, and (iii) an {\bf open-set analysis} to assess the false match probability when post-mortem samples are compared to live samples.

\subsection{Failed Comparisons Statistics}

\begin{table*}[!t]
\renewcommand{\arraystretch}{1.4}
\centering
\caption{Number of failed comparisons (genuine and impostor) and total number of comparisons calculated for all variants of analyses done in this study. Numbers are presented separately for each iris recognition method and suggest that iris image quality control mechanisms differ significantly among methods when post-mortem samples are used.}\vskip-1mm
\begin{tabular}[t]{p{3cm}|l|c|c|c|c}
\multicolumn{2}{l|}{} & \multicolumn{2}{c|}{\bf Short-term analysis} &\multicolumn{2}{c}  {\bf Long-term analysis}\\
\multicolumn{2}{l|}{} & \bf NIR images  &  \bf R images  & \bf NIR images & {\bf R images } \\\hline
\multicolumn{2}{l|}{\bf Total number of comparisons (genunine + impostor)}  & 39,621 & 76,245  & 245,904 & 447,882\\\hline
\multirow{5}{3cm}{\bf Number of failures to compute a comparison score}
 &{\bf IriCore} & 0 & 0 & 282 (0.11\%) & 0 \\\cline{2-6}
 &{\bf MIRLIN}  & 0 & 0 & 106,989 (43.51\%)  & 179,641 (40.11\%) \\\cline{2-6}
 & {\bf OSIRIS}  & 4,376 (11.04\%) & 779 (1.02\%) & 30,178 (12.27\%) & 55,484 (5.69\%) \\\cline{2-6}
 & {\bf OSIRIS (manual)}  & 0 & 0 &  0 & 0 \\\cline{2-6}
 &{\bf VeriEye}  & 0  & 0 & 2,538 (1.03\%) & 19,999 (4.47\%) \\
\end{tabular}
\label{tab:FTE}
\end{table*}

Iris recognition methods are often equipped with image quality control mechanisms that prevent  comparing samples with unacceptable properties. That is, if at least one of two samples being compared presents a less-than-accepted quality, the comparison fails and no valid comparison score is returned. 

Iris image quality seems to be defined differently in the implementations used in this study, since the numbers of failed comparisons is non-uniform across matchers. It is also difficult to hypothesize on what the exact reasons of these failures are, since out of four recognition algorithms, only one of them, IriCore, is explicitly said by the manufacturer to be compliant with the ISO/IEC 29794-6 standard on iris image quality. As for the open-source OSIRIS matcher, neither ISO/IEC metrics nor any other methods for discarding bad samples are implemented. For this method, samples denoted as those that failed to produce a template did so because of errors returned by low-level OpenCV routines used in OSIRIS, which may be a result of low quality of the processed samples. For the remaining two matchers, VeriEye and MIRLIN, there is no information how the iris image quality is assessed in these commercial products.

Table \ref{tab:FTE} summarizes numbers of all comparisons in short-term and long-term analyses along with numbers of the failed comparisons. For instance, the most restrictive method, MIRLIN, fails to compute almost 44\% of comparison scores when comparing NIR samples, and fails to compute over 40\% of comparison scores when comparing R images. In turn, the IriCore software was able to compute all short-term comparison scores and failed to compute only 0.11\% of long-term comparison scores for NIR images. Analyses presented in the remainder of this Section are based only on comparison scores that were correctly calculated. We should expect that methods controlling the iris image quality more restrictively should present a higher recognition accuracy (\eg lower EER) than the methods implementing weaker quality discrimination.

\subsection{Baseline Post-Mortem Performance: Short-Term Analysis}
\label{sec:analyses-desc}

\begin{figure*}[!htb]
	\centering
	\includegraphics[width=0.3\textwidth]{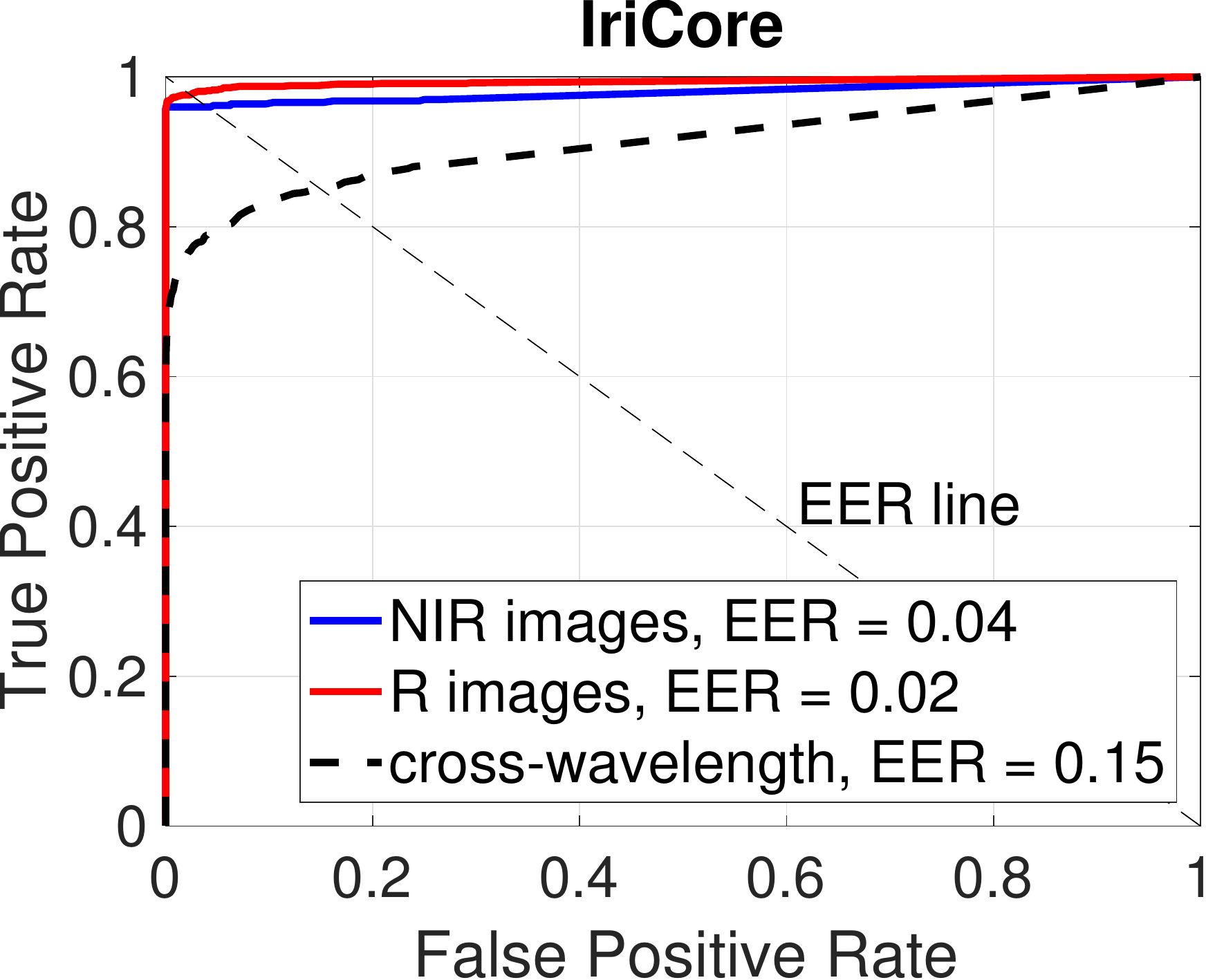}\hfill
	\includegraphics[width=0.3\textwidth]{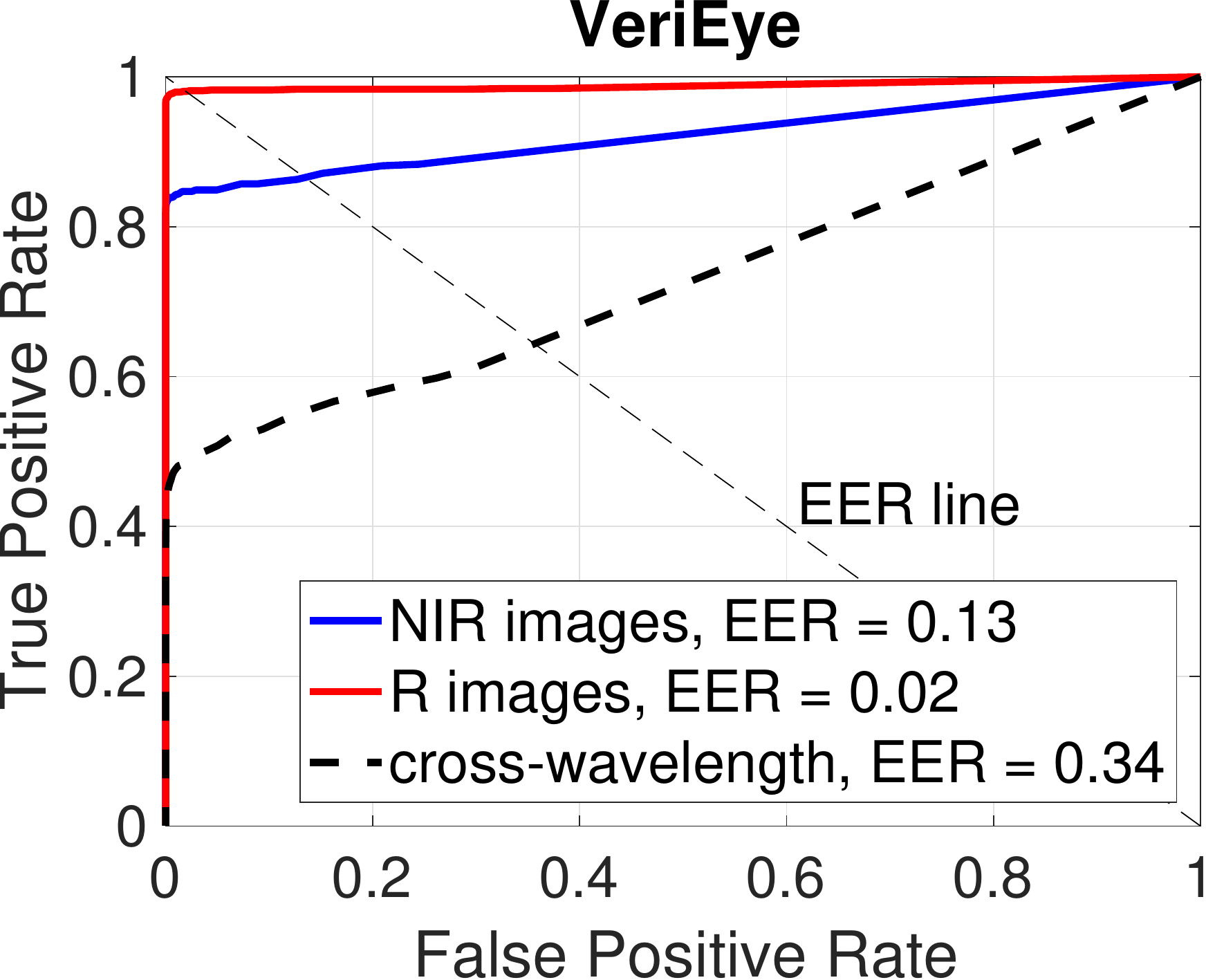}\hfill
	\includegraphics[width=0.3\textwidth]{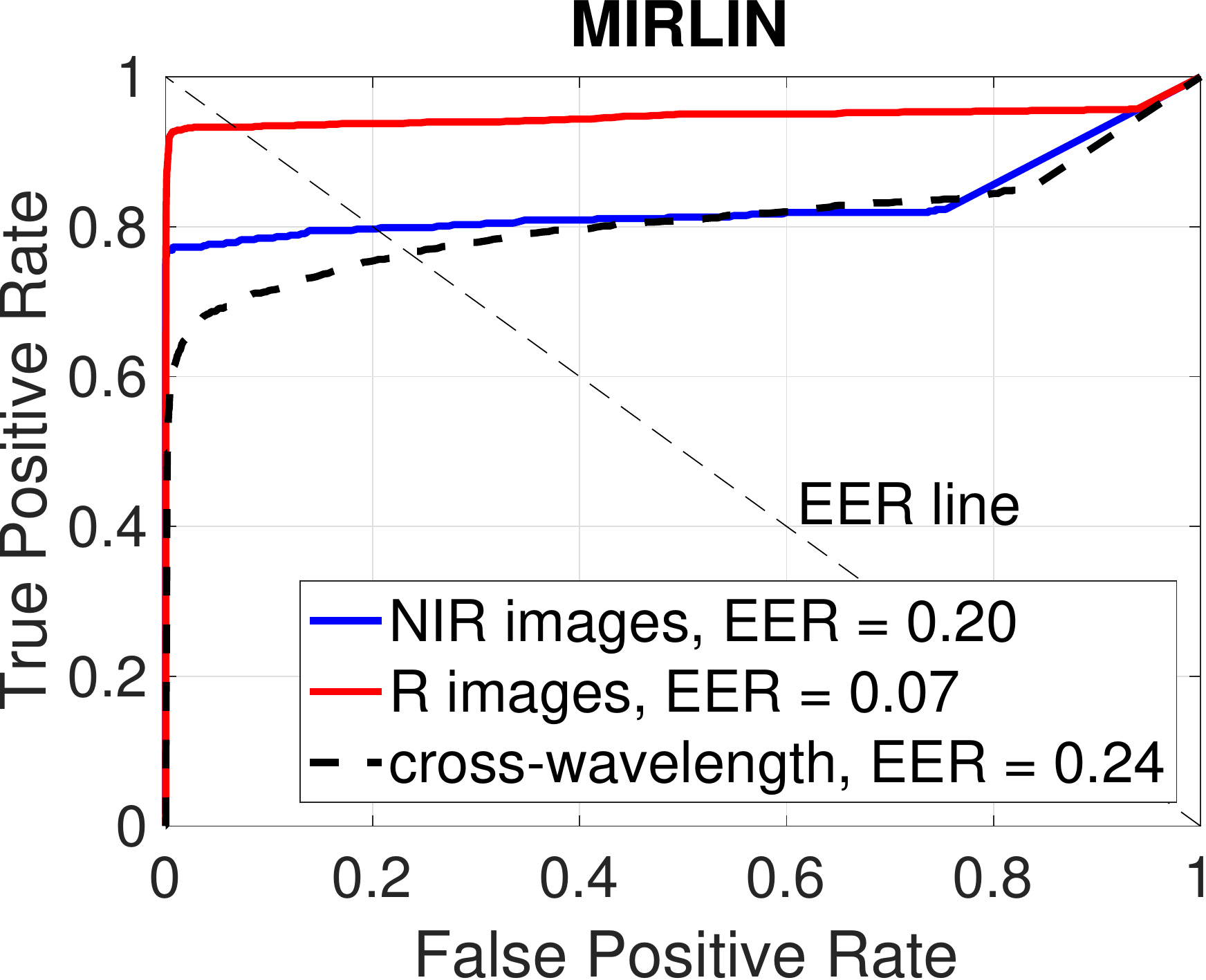}\\\vskip5mm
	\includegraphics[width=0.3\textwidth]{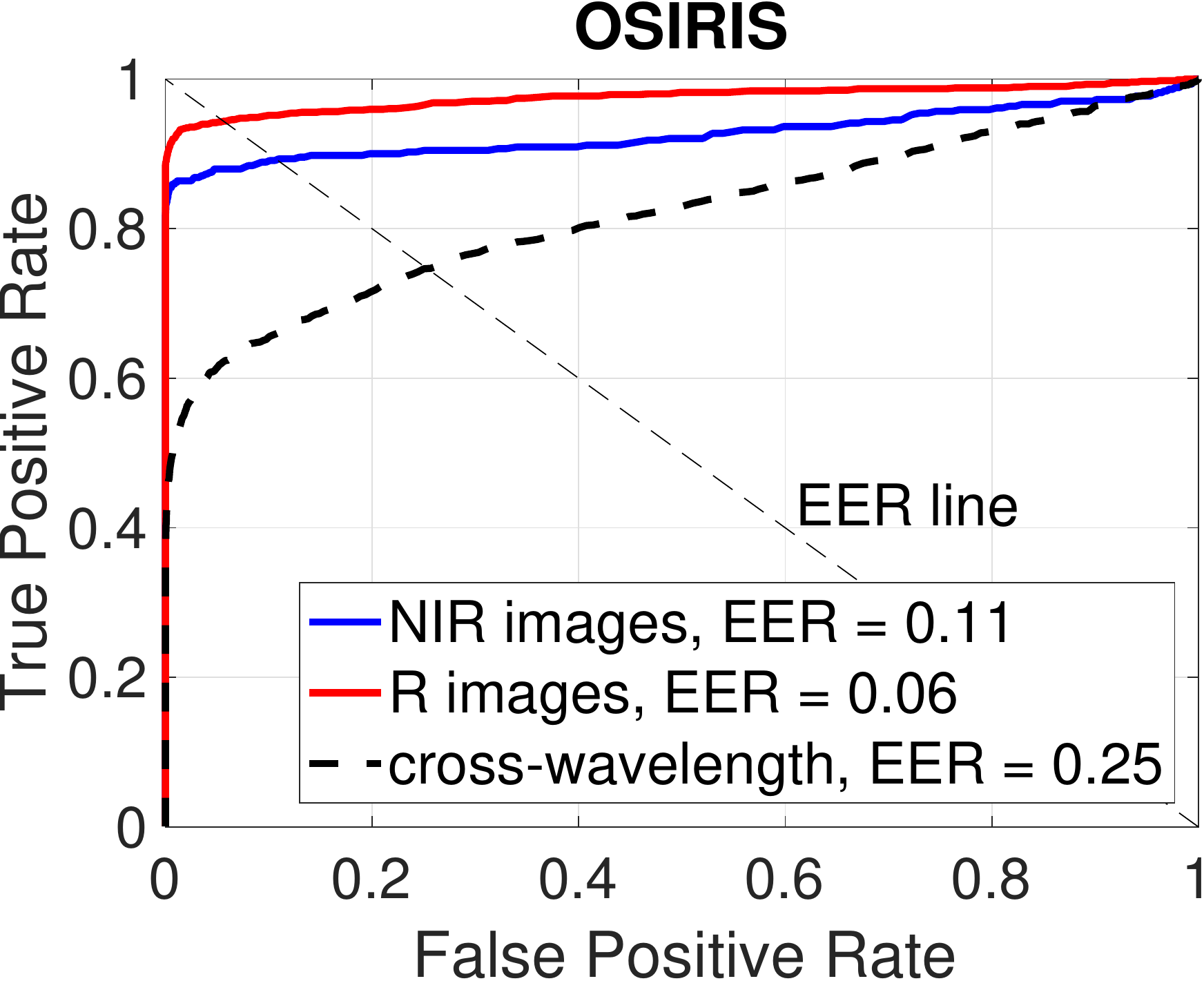}\hskip10mm
	\includegraphics[width=0.3\textwidth]{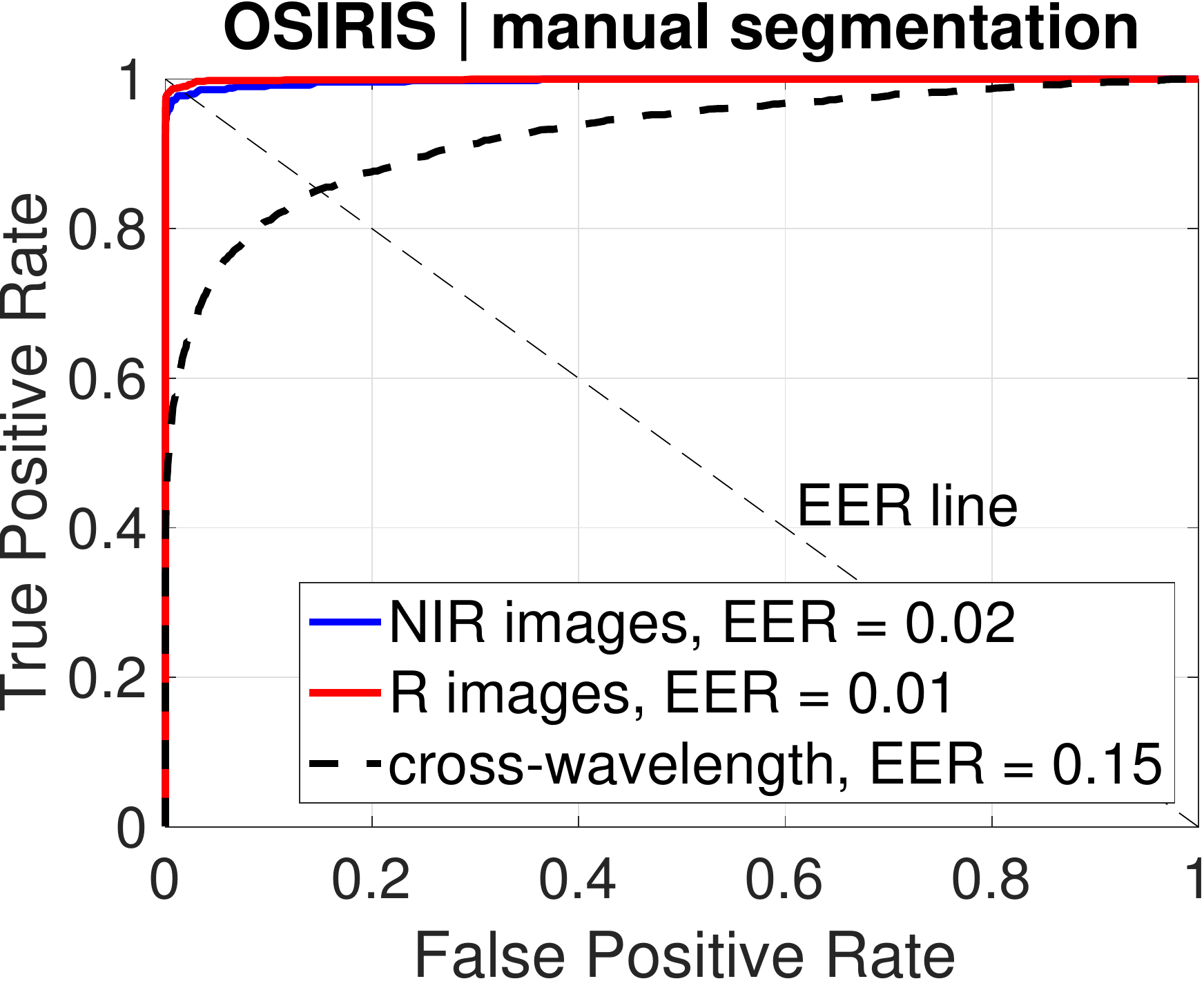}
	\caption{ROC curves for scores obtained when matching intra-session samples within the first acquisition session (\textcolor{myblue}{\bf NIR images: blue line}, \textcolor{myred}{\bf R images: red line}, \textbf{cross-wavelength matching: dashed black line}). Equal error rate (EER) is also shown for each case. Plots generated only for successful comparisons (\ie not rejected by the software due to low image quality).}
	\label{fig:shortTerm}
\end{figure*}

Fig. \ref{fig:shortTerm} illustrates Receiver Operating Characteristic (ROC) curves obtained in short-term analysis, when images were acquired within the first acquisition session (5-7 hours post-mortem). Iris recognition performs reasonably well on average in short-term post-mortem horizon, what conforms with earlier works on a smaller data sample \cite{TrokielewiczPostMortemICB2016,TrokielewiczPostMortemBTAS2016}. The best performing method, IriCore, achieved $EER=4\%$ for NIR images, and $EER=2\%$ for R images.

Interestingly, having taken iris images in visible light with a high-resolution camera, and using their red channel, allows for a better performance of all of the involved iris recognition methods when compared to the performance achieved for VGA (\ie low-resolution) NIR images. We can observe even six-fold gains in recognition accuracy for the VeriEye matcher, measured EER-wise (2\% error with R images versus 13\% error for NIR images). One of the contributing factors for such a favorable performance of R images is likely a dominant proportion of lightly-colored eyes in the dataset, cf. Sec. \ref{sec:DatabaseStats}, for which iris recognition is known to work well even without a specialized NIR camera, provided that the images are of good quality \cite{TrokielewiczBartuziJTIT}. For completeness, we also include an analysis of cross-spectral matching scenarios, where R images are matched against NIR images (dashed black lines in Fig. \ref{fig:shortTerm}), however, the results of such matching are discouraging enough not to be investigated any further, and thus are not considered in the long-term analysis. Apparently, for post-mortem images, the differences in sample presentation under different illumination are large enough to make such a scenario unusable.

Note that all algorithms, except for the OSIRIS, did not reject any sample due to low quality (cf. `Short-term analysis' columns in Tab. \ref{tab:FTE}), which makes these methods perfectly viable for samples collected a few hours after death. The worst method in short-term analysis, MIRLIN, presents equal error rate of 7\% for R images and 20\% for NIR images. However, this could possibly be compensated by more restrictive quality control since MIRLIN, as IriCore, did not reject any sample prior to matching. Even so, the worst result (EER=20\%) is still far better than 50\% expected for a random classifier, indicating a possibility to correctly recognize a large subset of cadaver irises.

Finally, when examining results obtained when matching manually segmented samples using the OSIRIS matcher (bottom-right plot in Fig. \ref{fig:shortTerm}), we may draw a conclusion that most of the recognition errors can be associated with erroneous execution of the automatic segmentation stage, as this method, when processing manually segmented images, presents $EER=1\%$ for R images and $2\%$ for NIR images. This shows that when post-mortem samples are carefully segmented to represent the correct iris region, iris recognition is a viable method for identification.

\subsection{Biometric Capability Decay Study: Long-Term Analysis}

\begin{figure*}[!t]
	\centering
	\includegraphics[width=0.49\textwidth]{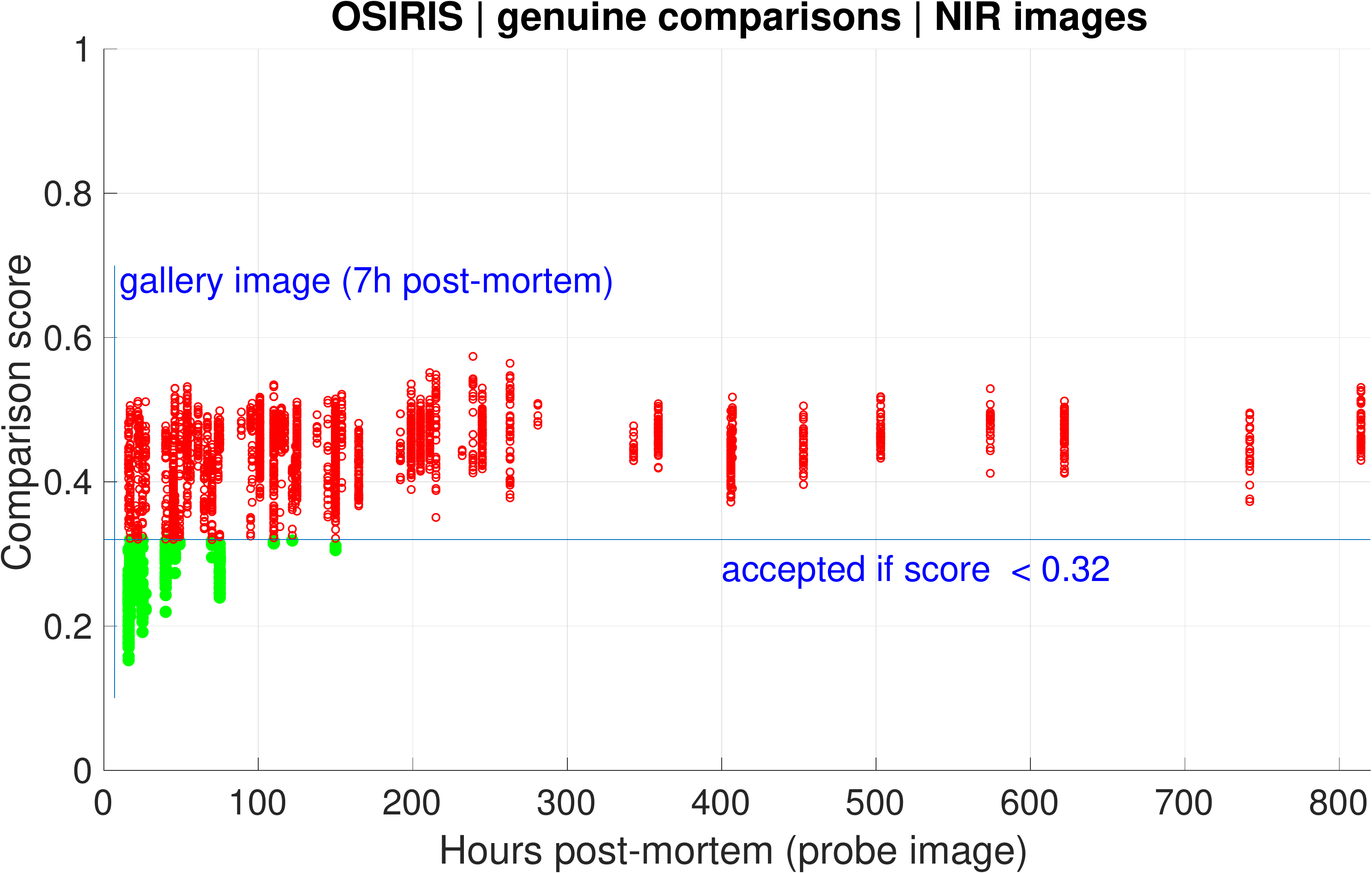}\hfill
	\includegraphics[width=0.49\textwidth]{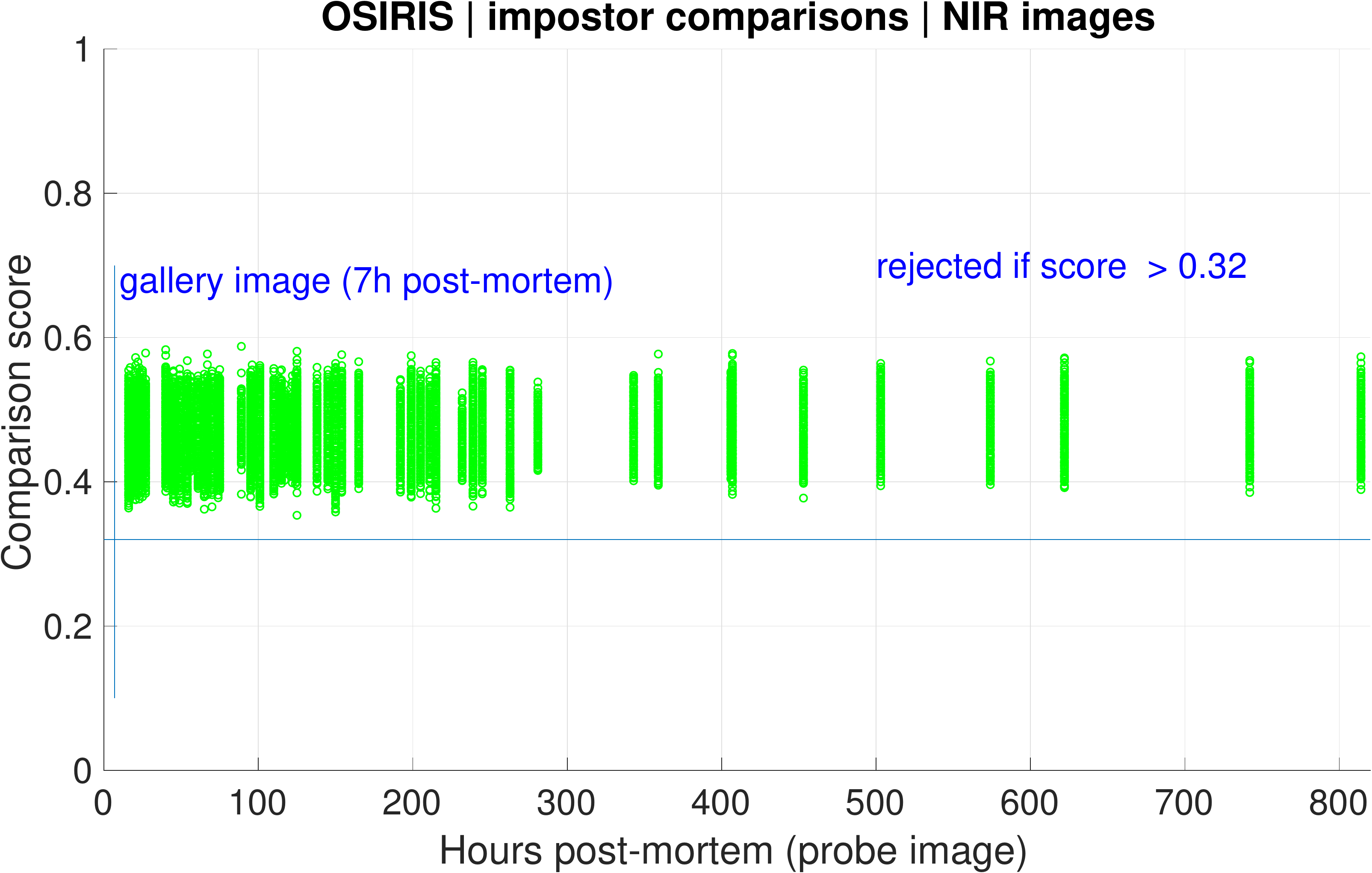}\\\vskip5mm
	\includegraphics[width=0.49\textwidth]{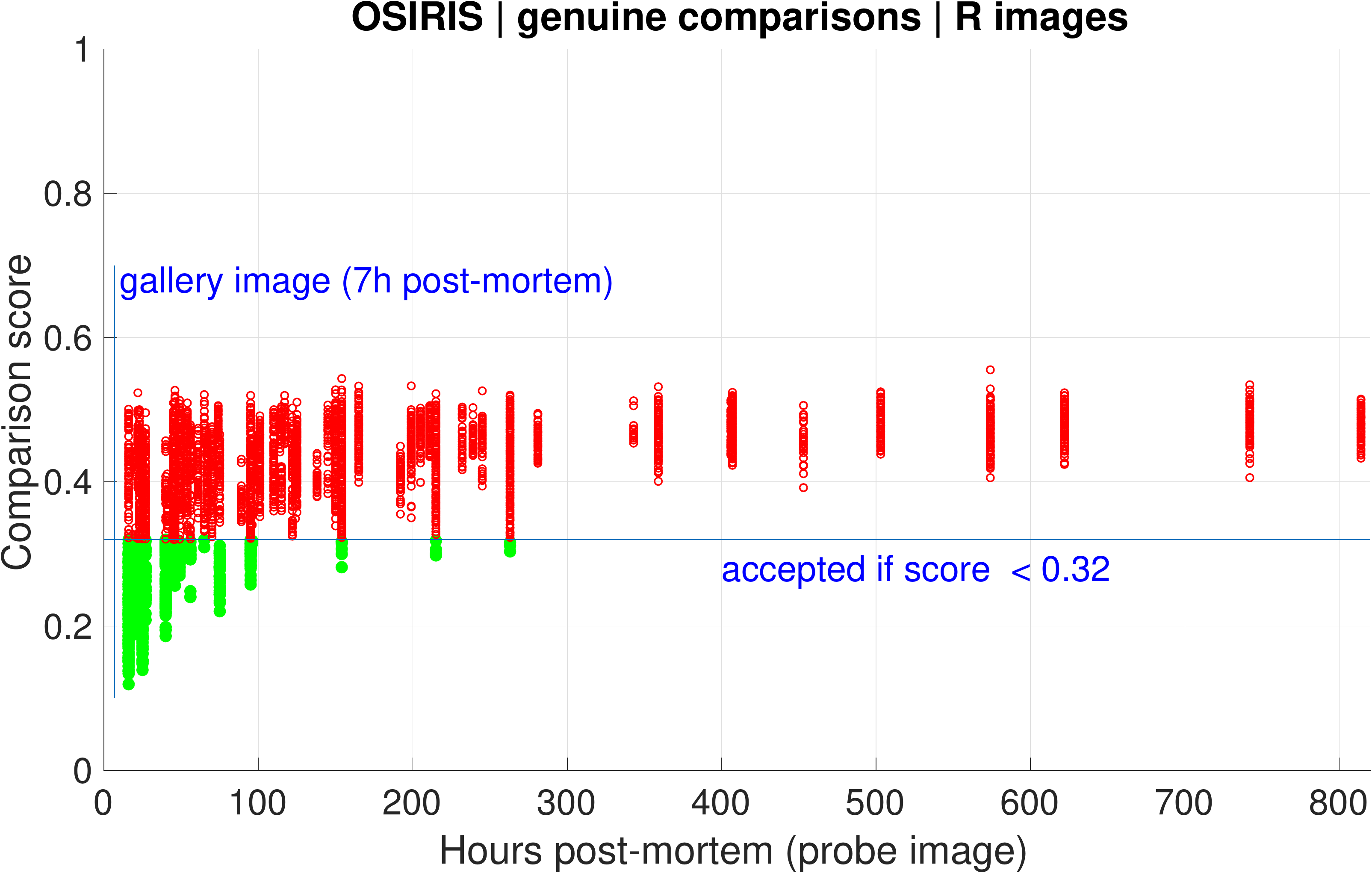}\hfill
	\includegraphics[width=0.49\textwidth]{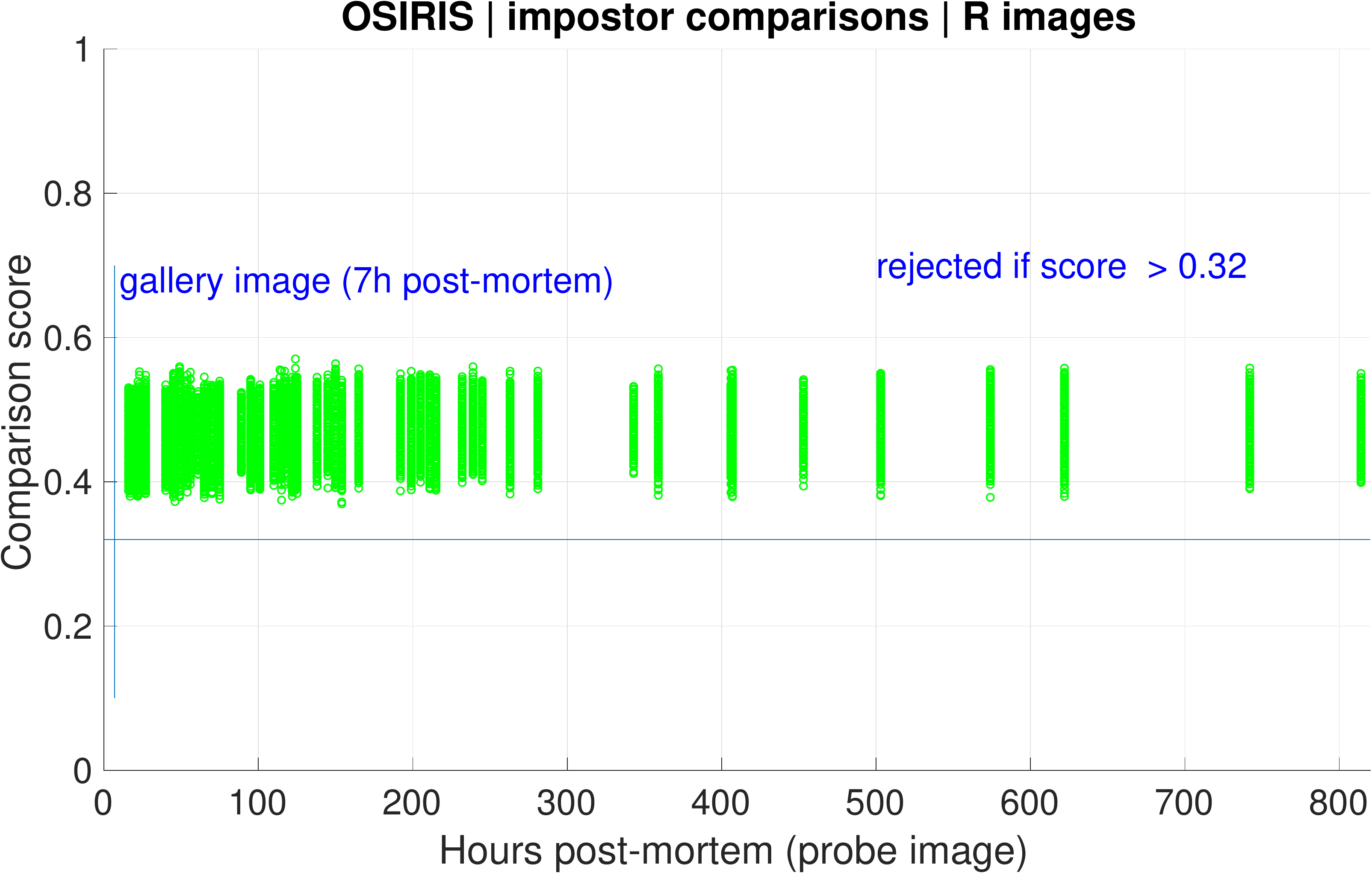}\\	
	\caption{Long-term analysis for \textbf{all subjects} for the \textbf{OSIRIS} matcher using automatic iris segmentation, for both NIR and R samples. \textcolor{mygreen}{\bf Green dots represent correct behavior, \ie a match for genuine pair, and a non-match for impostor pair}, while \textcolor{myred}{\bf red dots correspond to incorrect behavior, \ie a false non-match for a genuine pair, and a false match for impostor pair} between samples acquired in the first session (after 5-7 hours) and samples acquired in the following sessions.}
	\label{fig:longTerm_OS}
\end{figure*}

\begin{figure}[!htb]
	\centering
	\begin{subfigure}[!htb]{0.24\textwidth}
	\centering
		\includegraphics[width=0.49\textwidth]{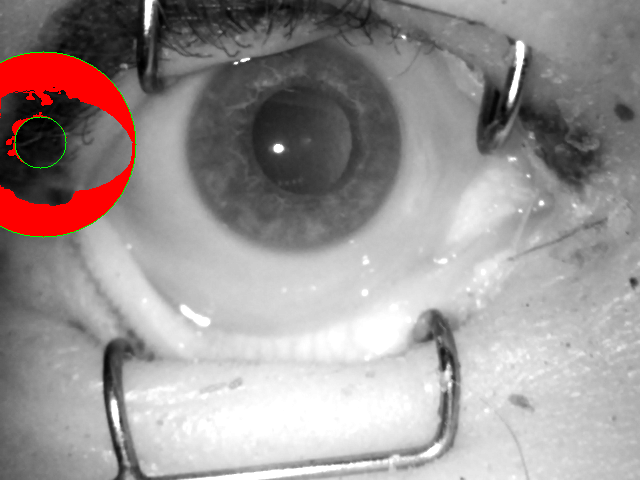}\hskip0.5mm
		\includegraphics[width=0.49\textwidth]{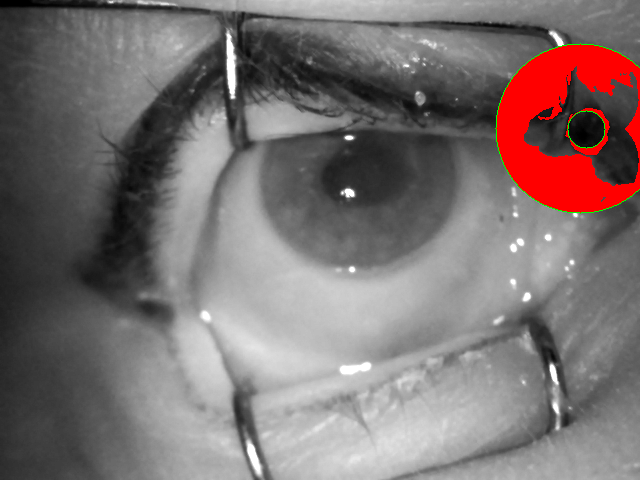}
        \caption{False non-match (0.60)}
    	\end{subfigure}\hskip1.5mm
	\begin{subfigure}[!htb]{0.24\textwidth}
	\centering
		\includegraphics[width=0.49\textwidth]{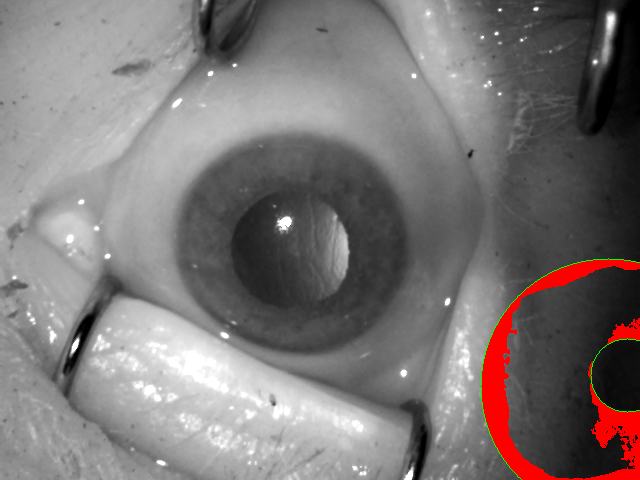}\hskip0.5mm
		\includegraphics[width=0.49\textwidth]{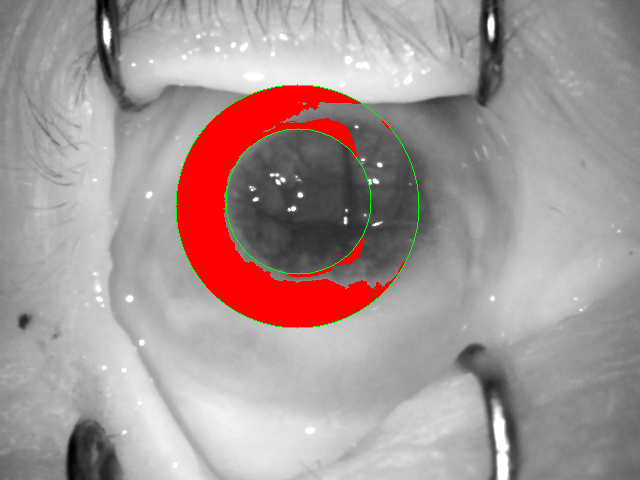}
        \caption{False match (0.08)}
    	\end{subfigure}
	\caption{Example image pairs generating a false non-match and a false match for the {\bf OSIRIS} matcher {\bf prior to score normalization}. The comparison scores are shown in brackets. Left image of each pair is the gallery sample. We assume that OSIRIS returns a match when the comparison score is below 0.32. After score normalization, the false match shown above is rectified.}
	\label{fig:FM-FNM-OS}
\end{figure}

\begin{figure*}[!tb]
	\centering
	\includegraphics[width=0.49\textwidth]{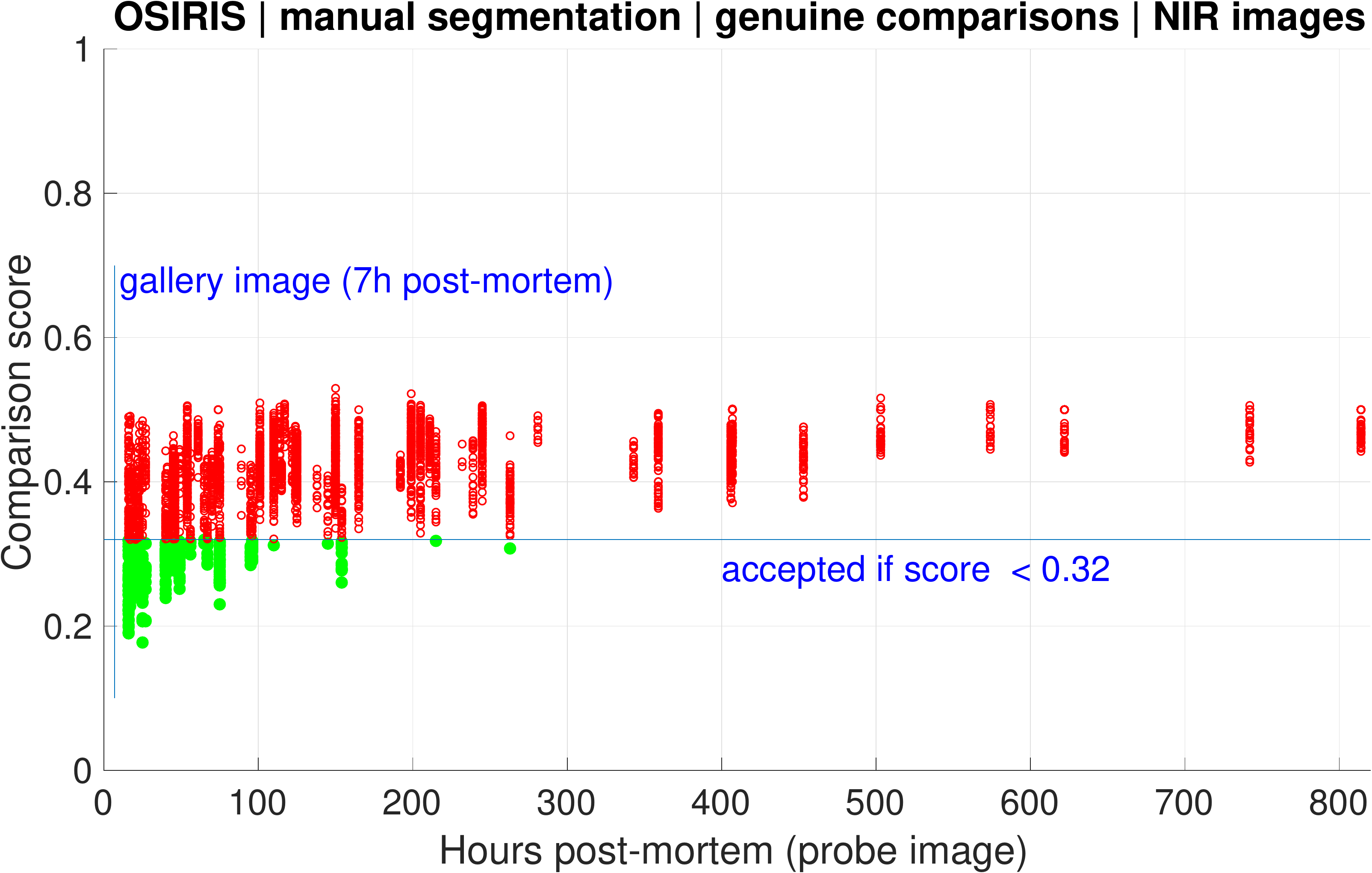}\hfill
	\includegraphics[width=0.49\textwidth]{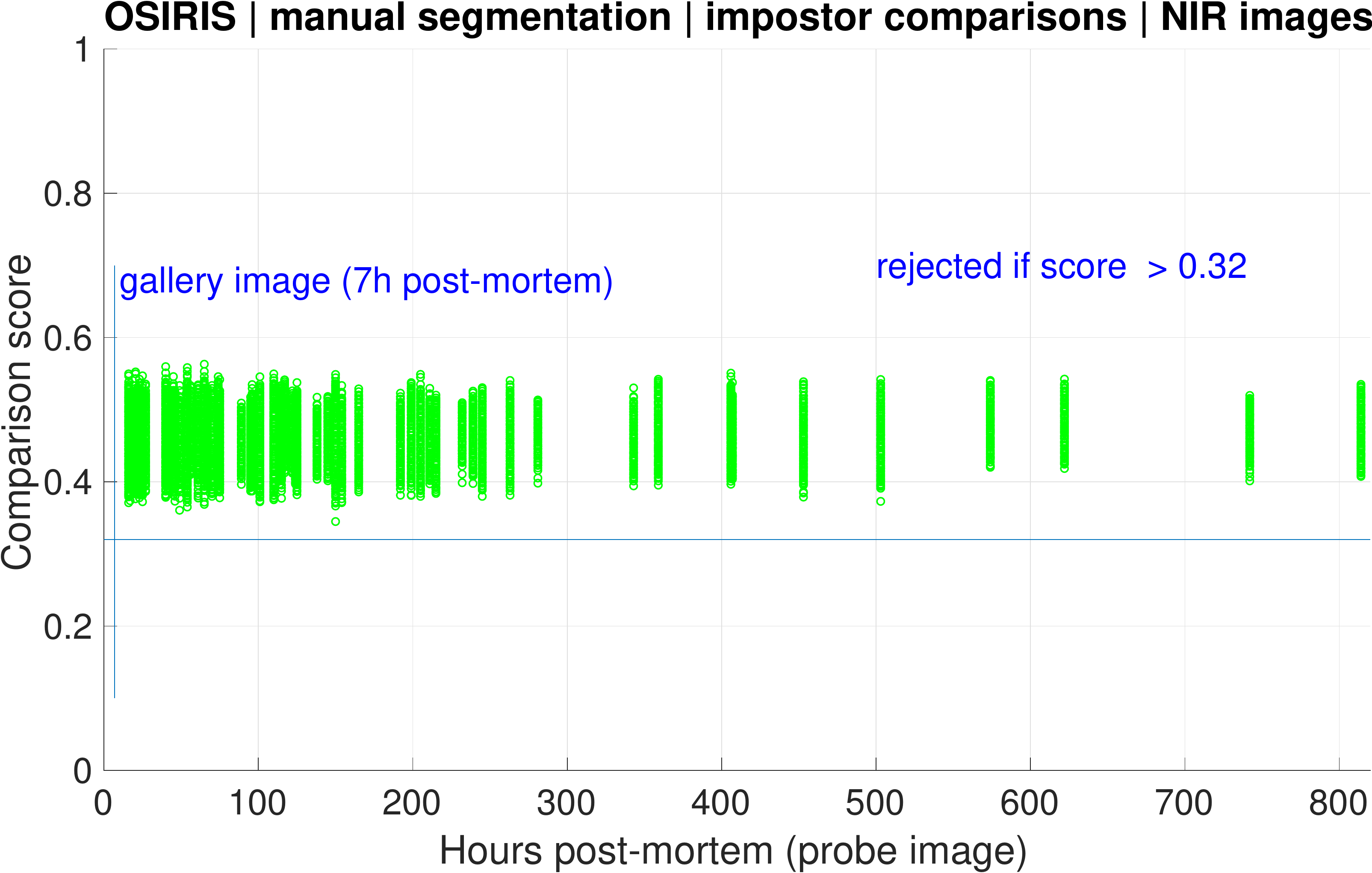}\\\vskip5mm
	\includegraphics[width=0.49\textwidth]{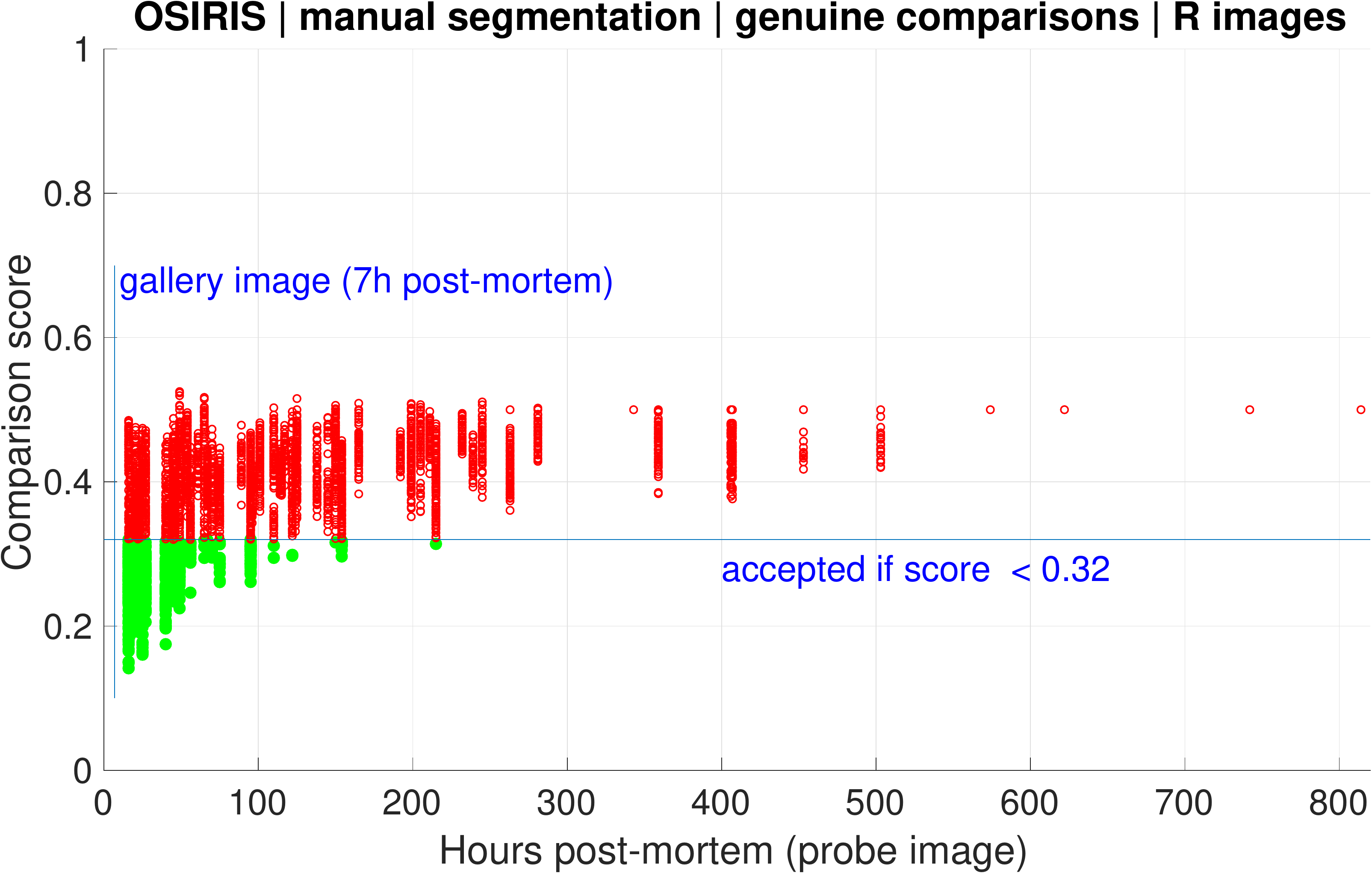}\hfill
	\includegraphics[width=0.49\textwidth]{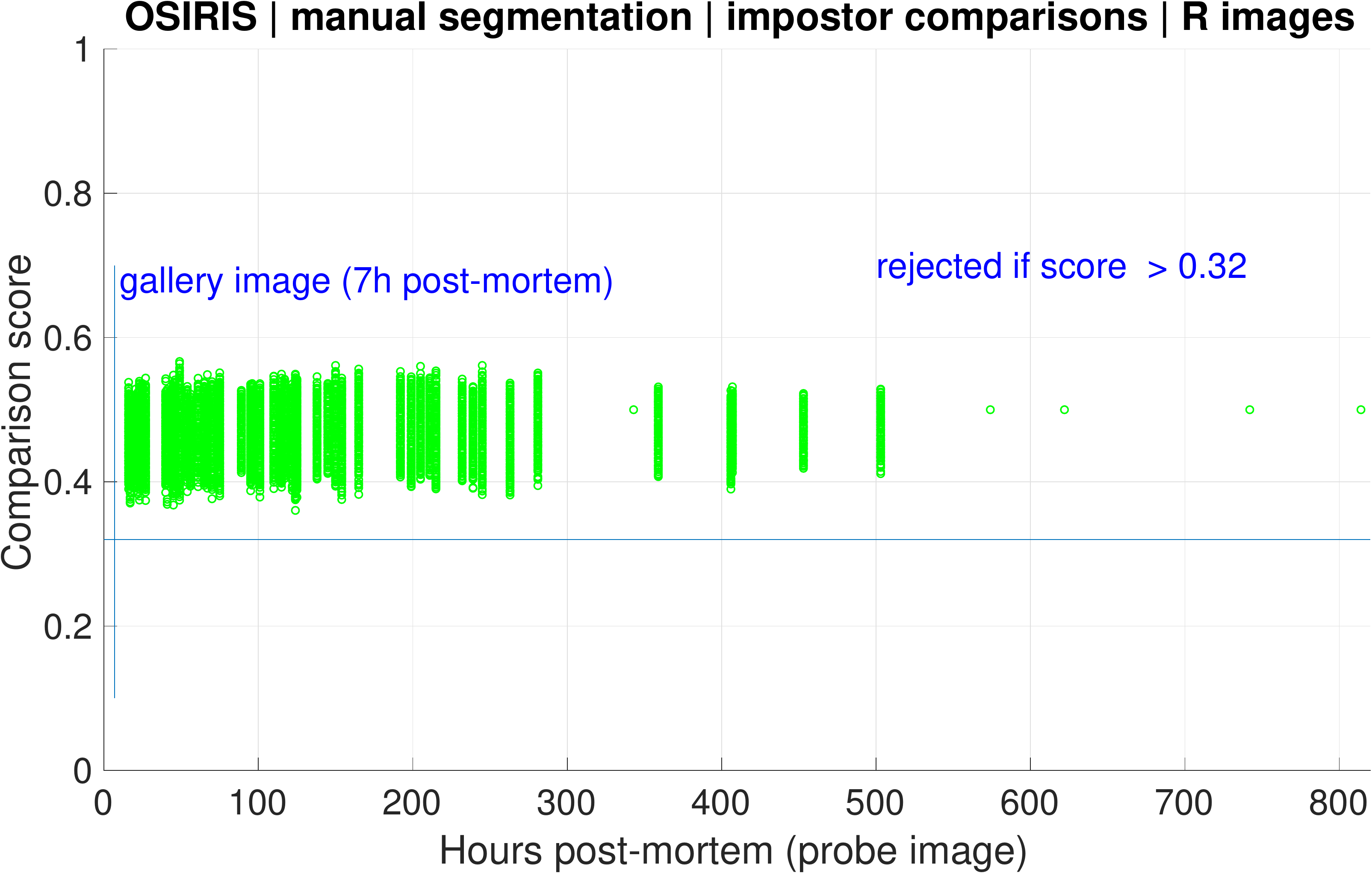}\\	
	\caption{Same as in Fig. \ref{fig:longTerm_OS}, but with manually corrected image segmentation.}
	\label{fig:longTerm_OSm}
\end{figure*}

\begin{figure*}[!htb]
	\centering
	\includegraphics[width=0.49\textwidth]{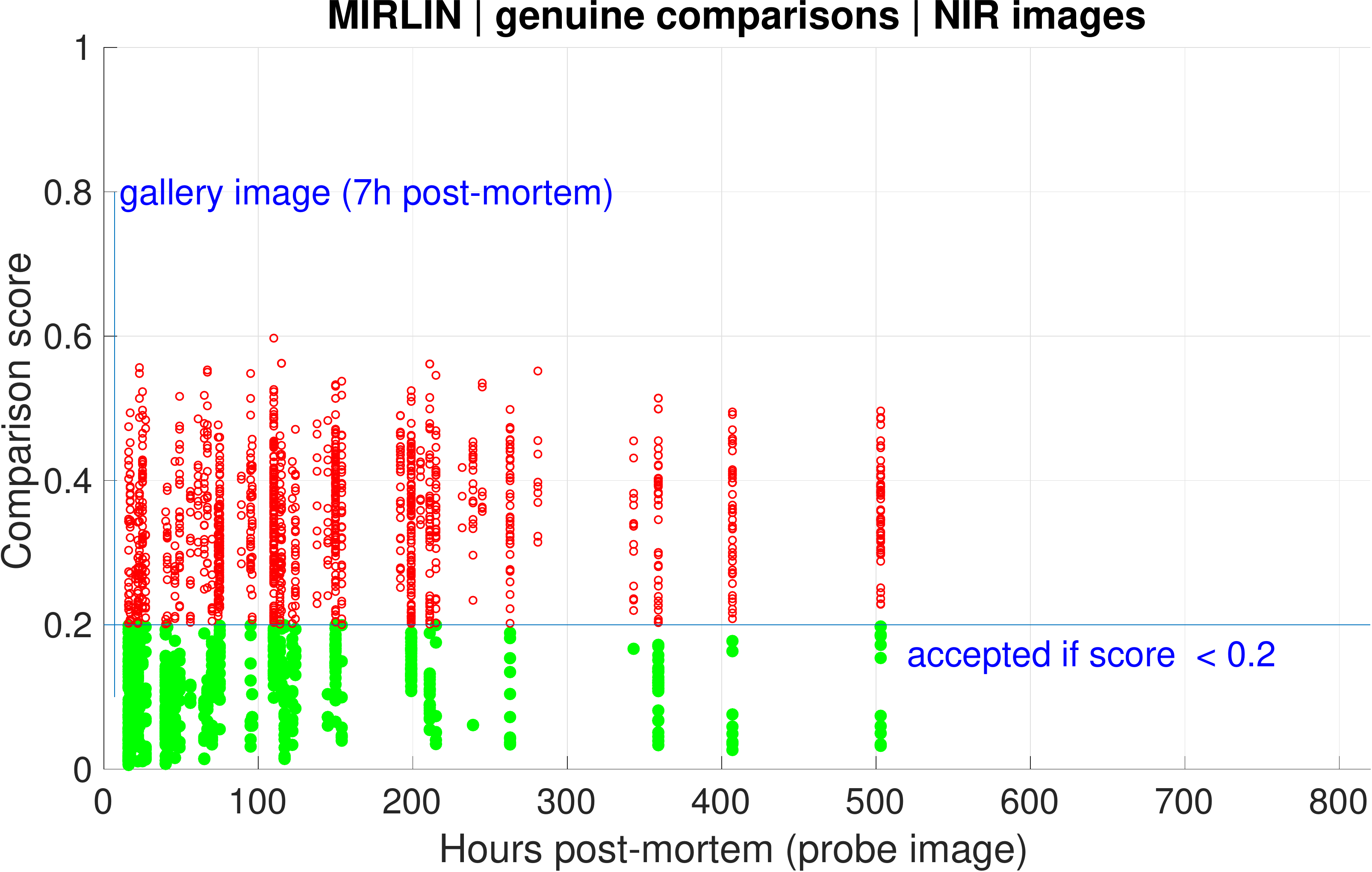}\hfill
	\includegraphics[width=0.49\textwidth]{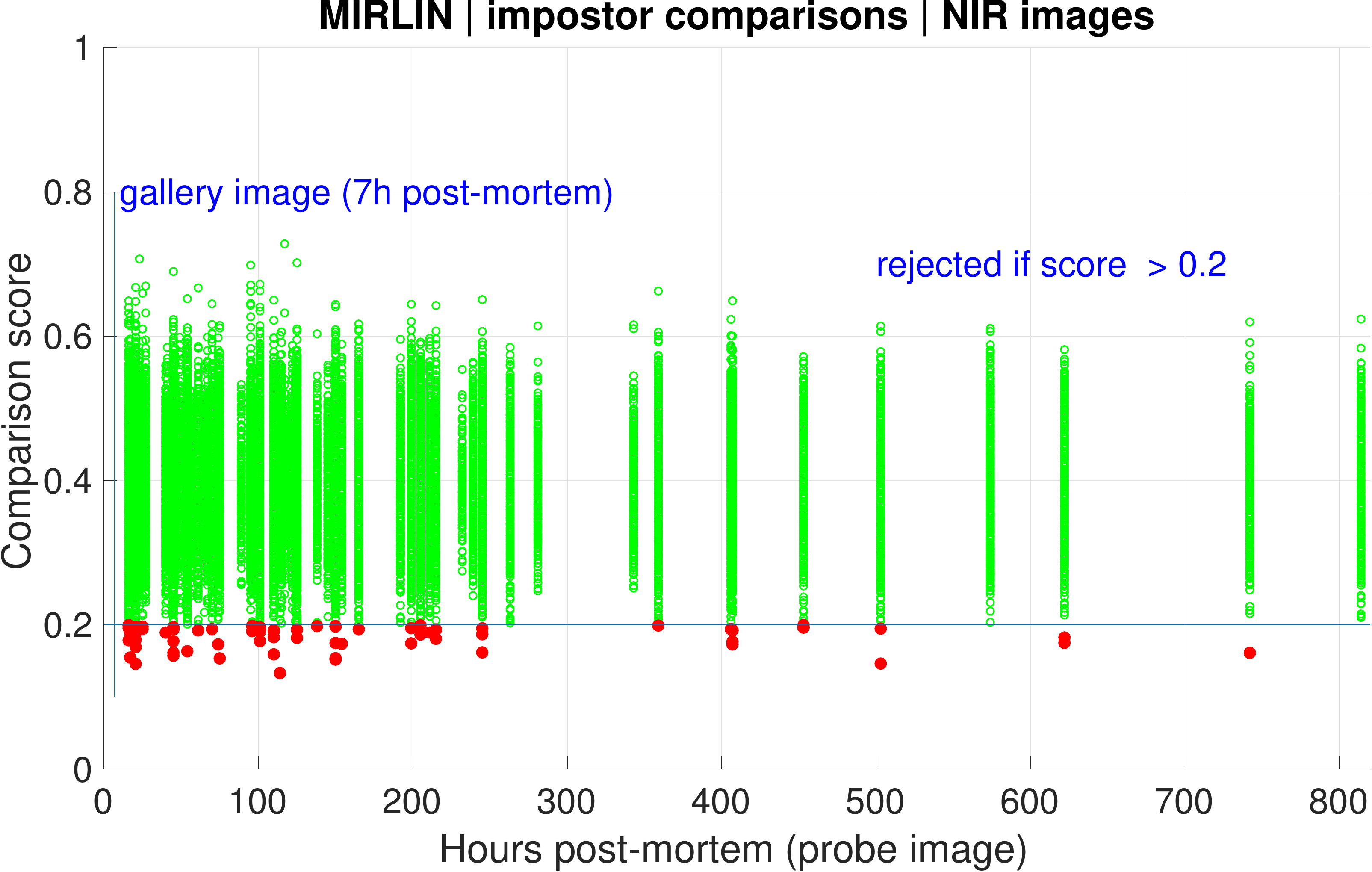}\\\vskip5mm
	\includegraphics[width=0.49\textwidth]{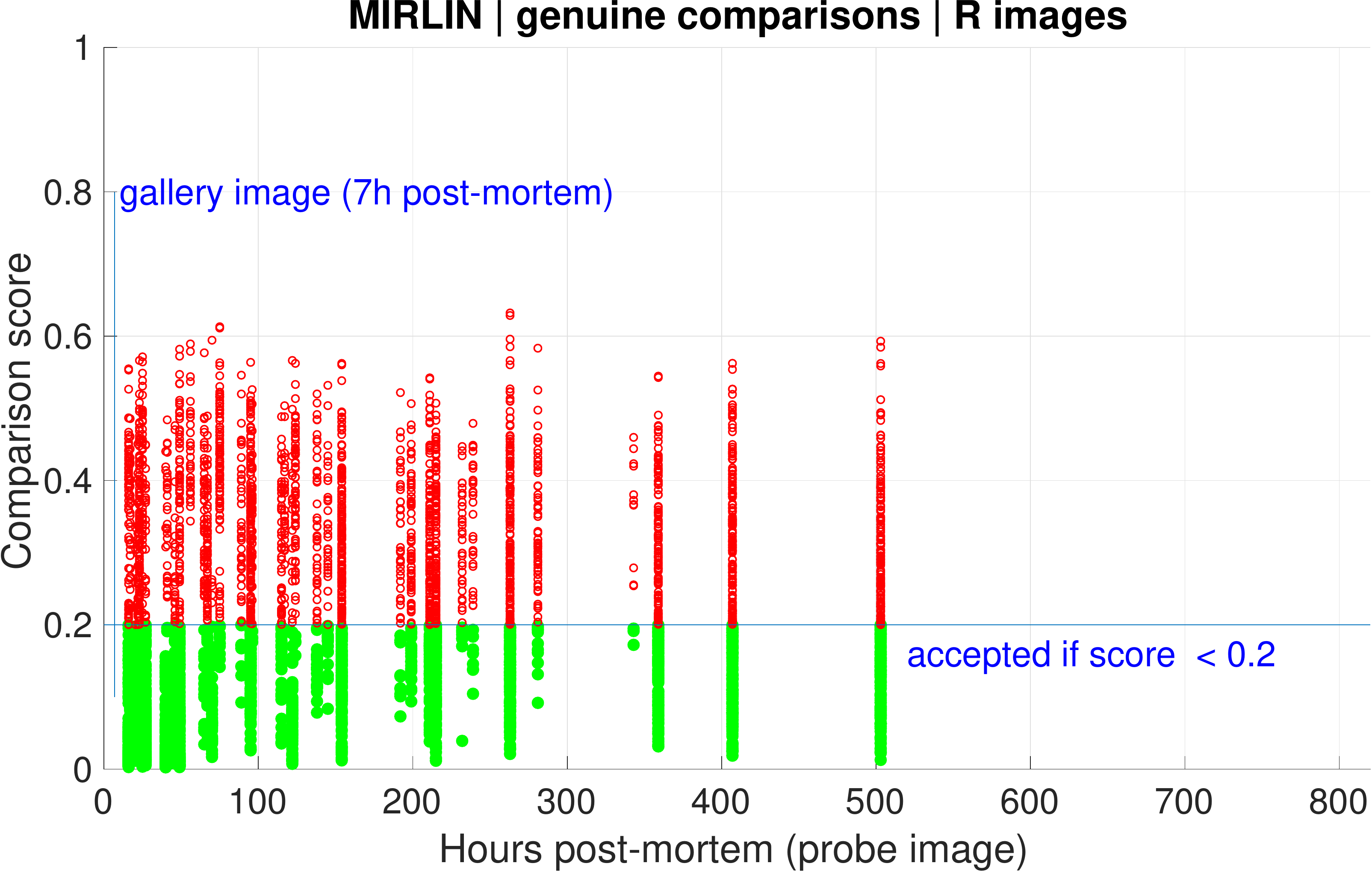}\hfill
	\includegraphics[width=0.49\textwidth]{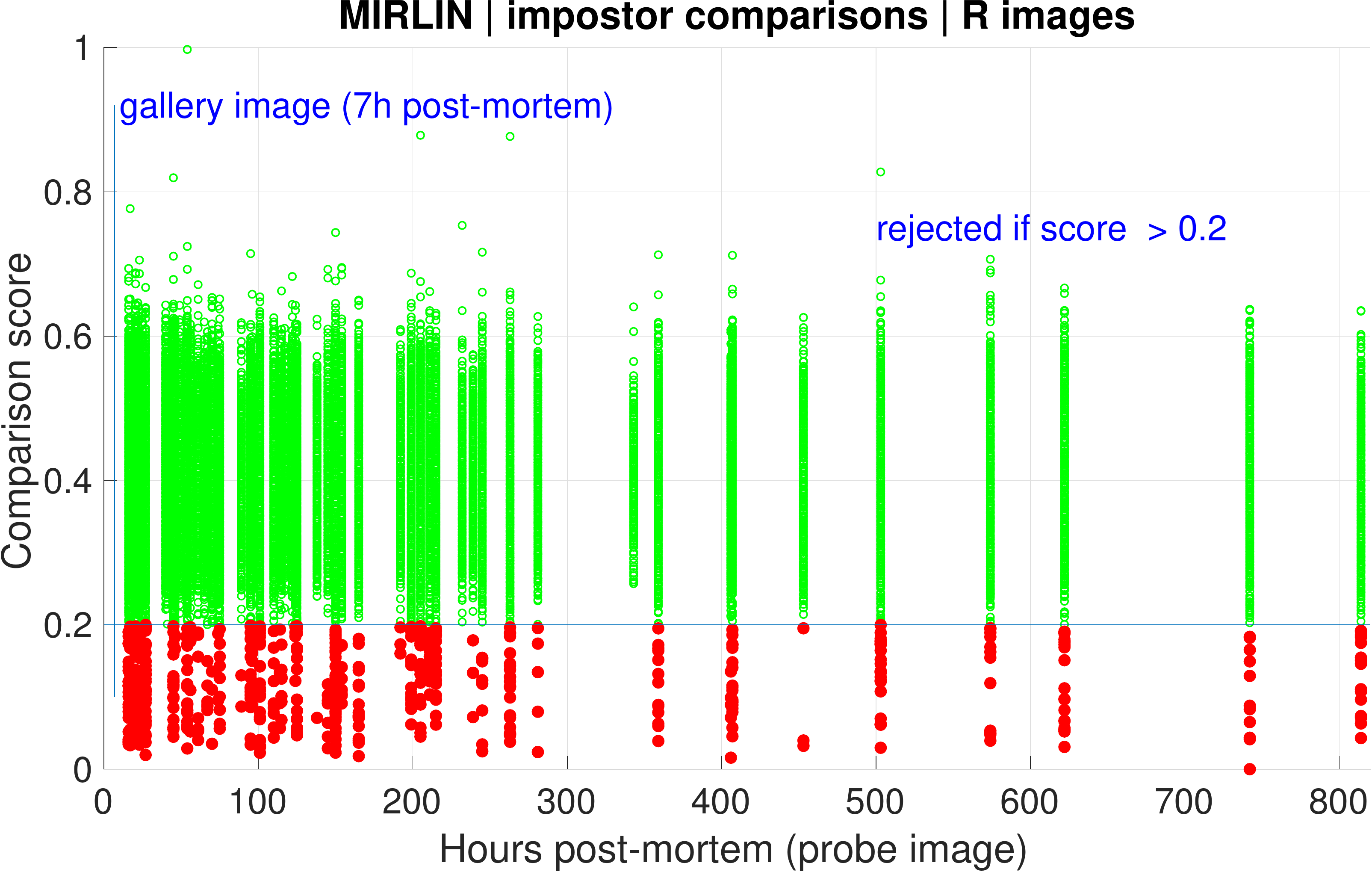}\\	
	\caption{Same as in Fig. \ref{fig:longTerm_OS}, but for the \textbf{MIRLIN} matcher.}
	\label{fig:longTerm_ML}
\end{figure*}

\begin{figure}[!htb]
	\centering
	\begin{subfigure}[!htb]{0.24\textwidth}
	\centering
		\includegraphics[width=0.49\textwidth]{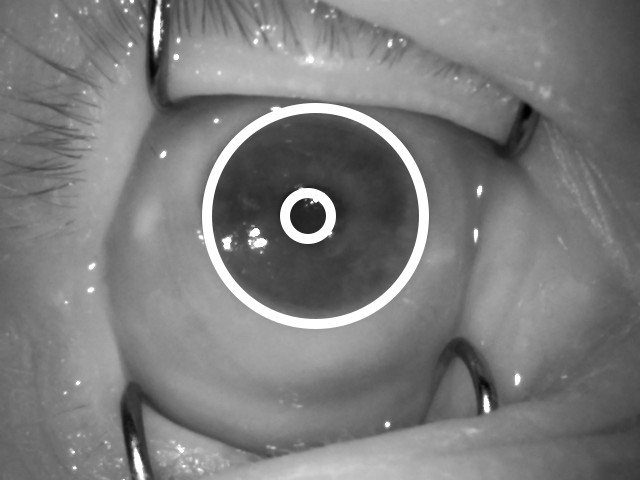}\hskip0.5mm
		\includegraphics[width=0.49\textwidth]{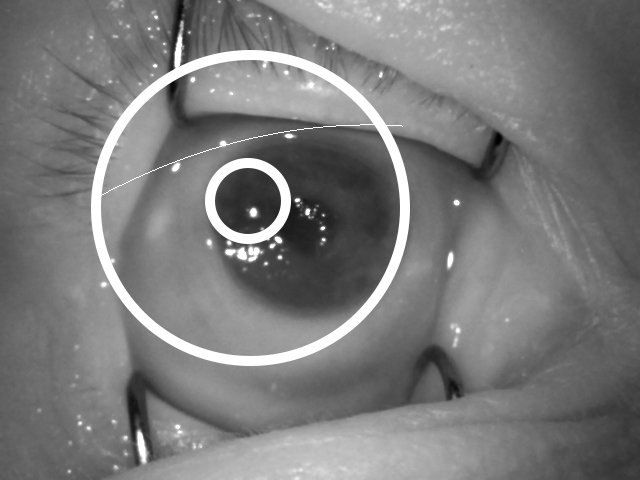}
        \caption{False non-match (0.62)}
    	\end{subfigure}\hskip1.5mm
	\begin{subfigure}[!htb]{0.24\textwidth}
	\centering
		\includegraphics[width=0.49\textwidth]{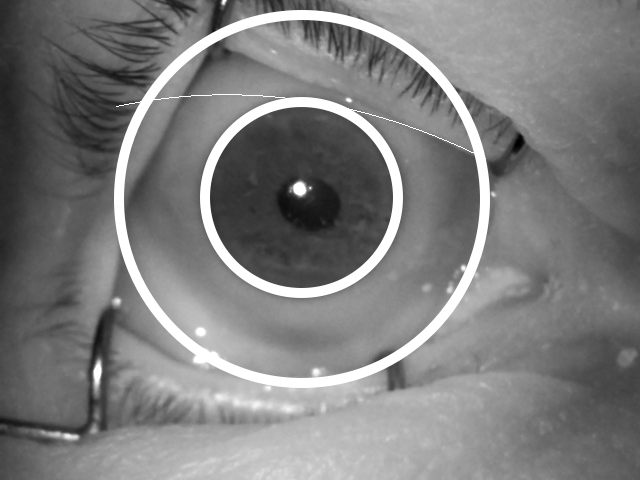}\hskip0.5mm
		\includegraphics[width=0.49\textwidth]{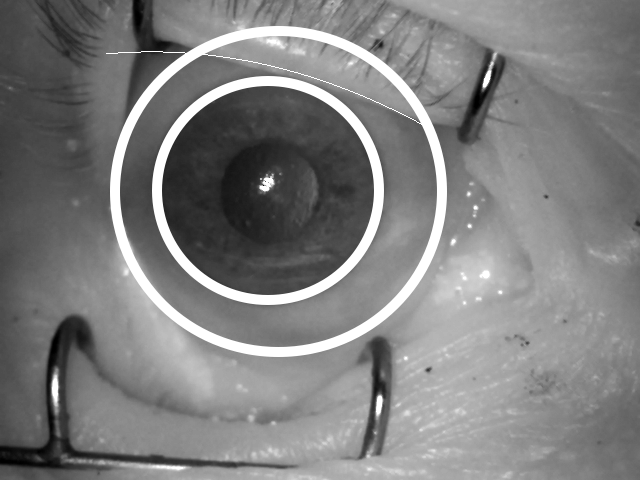}
        \caption{False match (0.09)}
    	\end{subfigure}
	\caption{Same as in Fig. \ref{fig:FM-FNM-OS}, except for the {\bf MIRLIN} matcher. We assume that MIRLIN returns a match when the comparison score is below 0.2.}
	\label{fig:FM-FNM-ML}
\end{figure}

\begin{figure*}[!htb]
	\centering
	\includegraphics[width=0.49\textwidth]{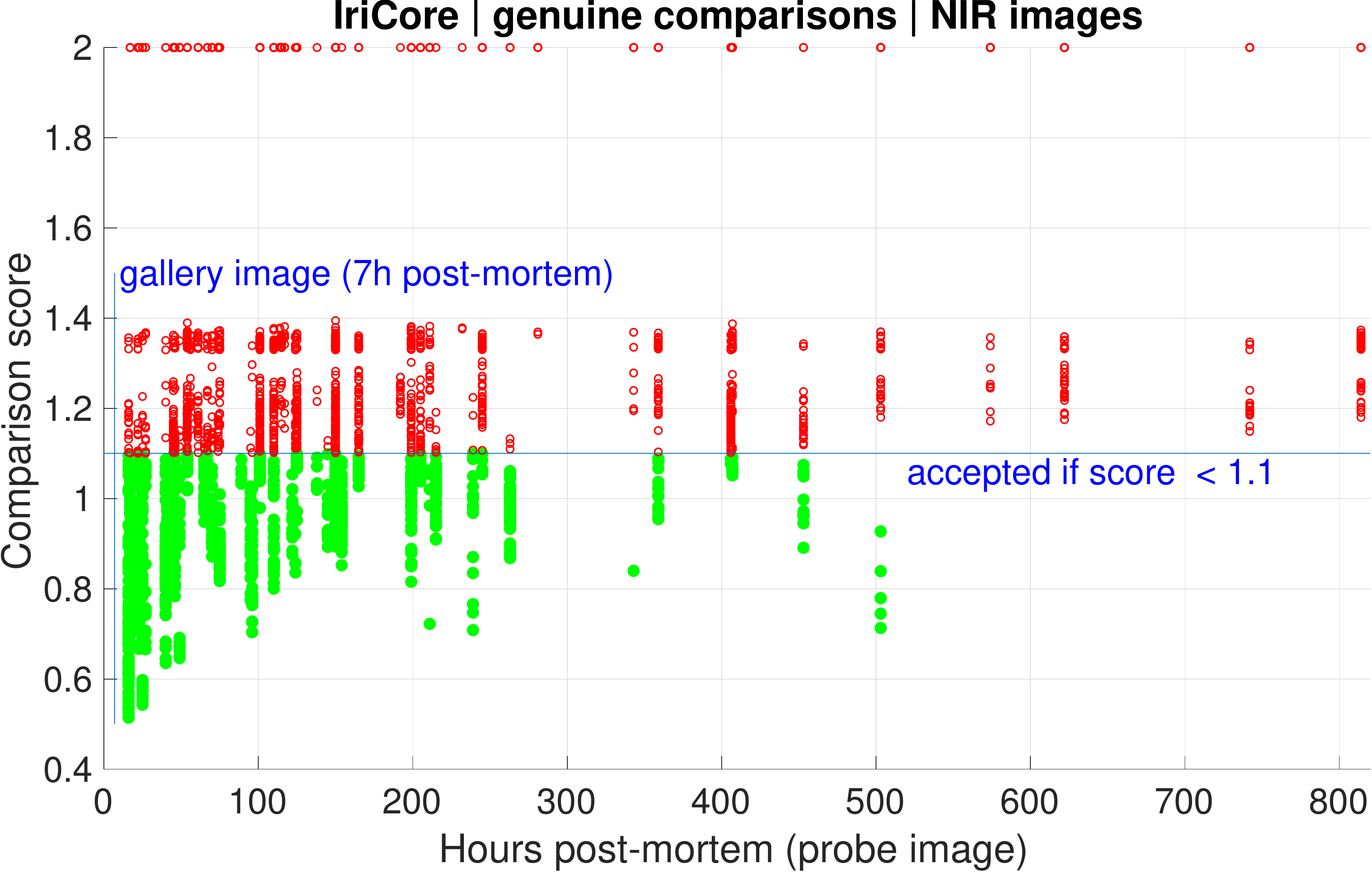}\hfill
	\includegraphics[width=0.49\textwidth]{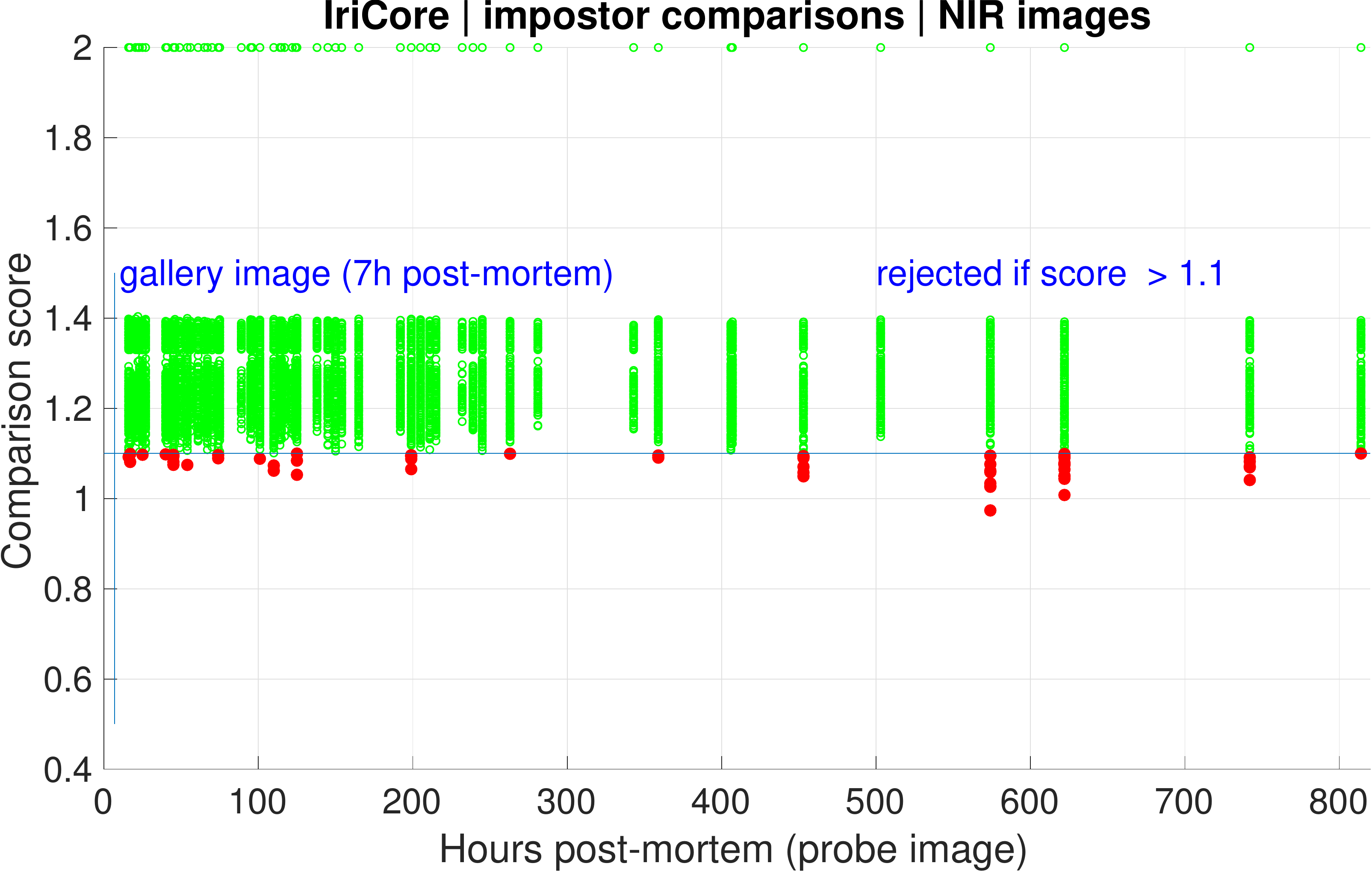}\\\vskip5mm
	\includegraphics[width=0.49\textwidth]{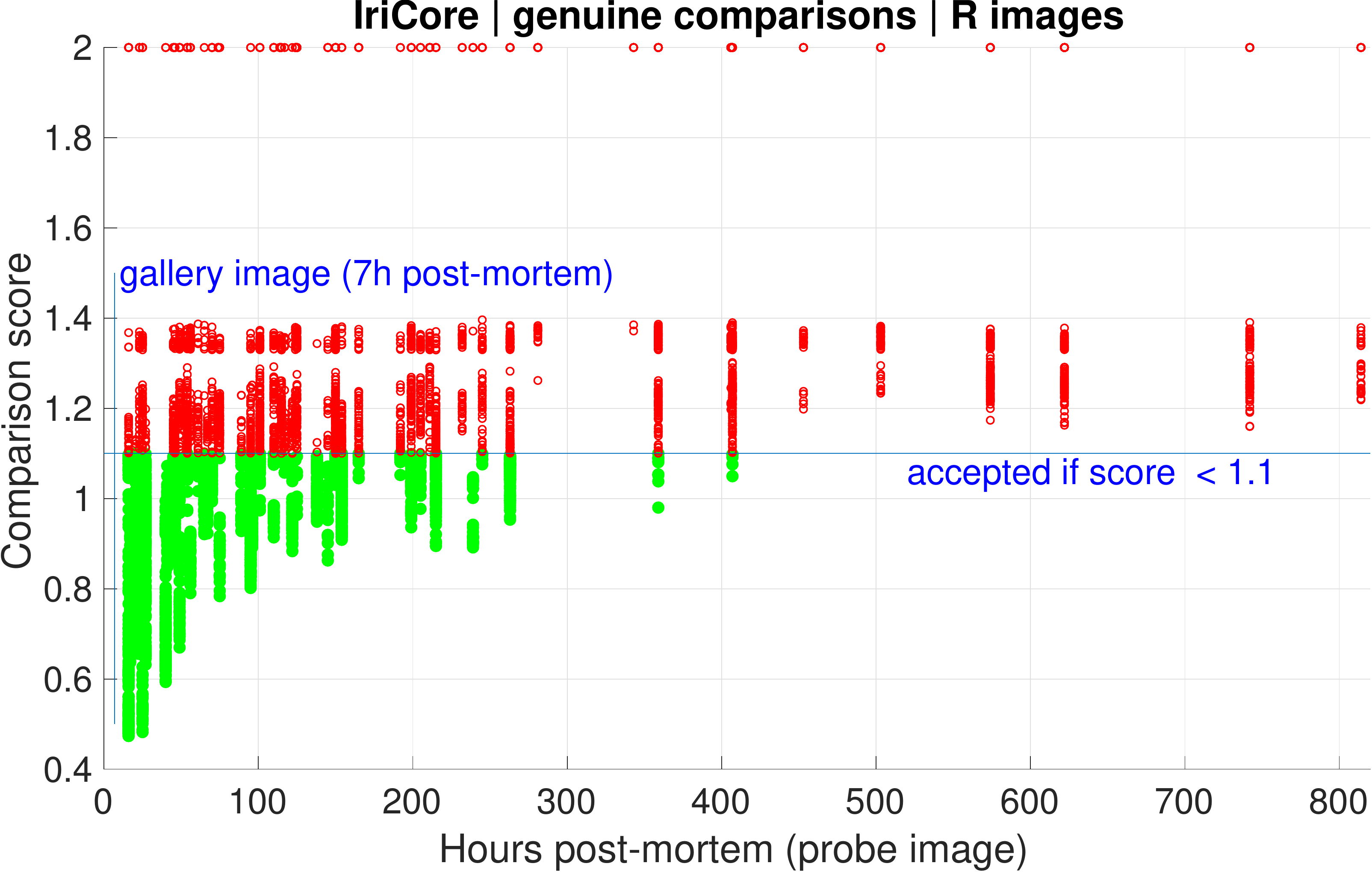}\hfill
	\includegraphics[width=0.49\textwidth]{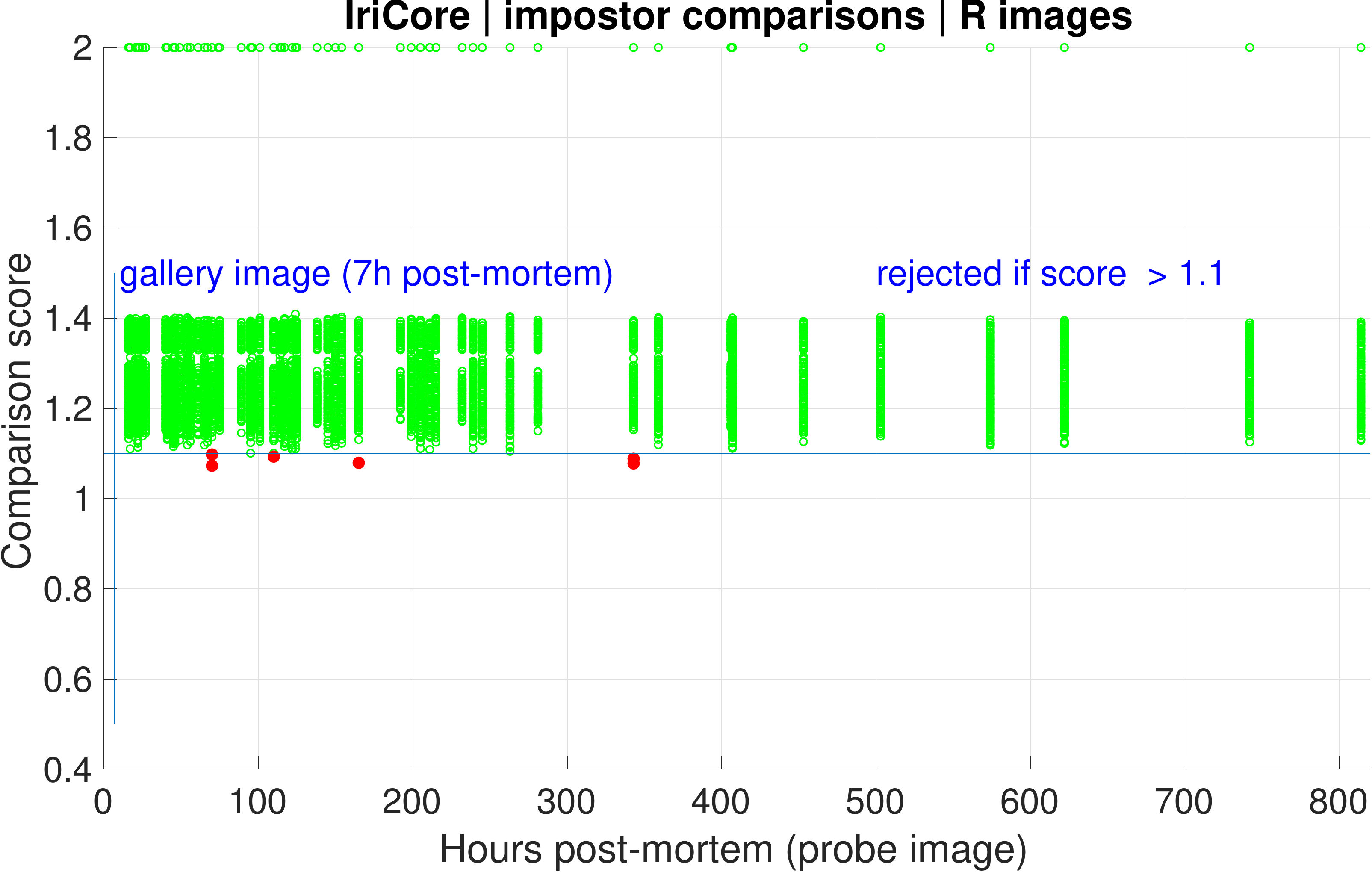}\\	
	\caption{Same as in Fig. \ref{fig:longTerm_OS}, but for the \textbf{IriCore} matcher.}
	\label{fig:longTerm_IC}
\end{figure*}

\begin{figure}[!htb]
	\centering
	\begin{subfigure}[!htb]{0.24\textwidth}
	\centering
		\includegraphics[width=0.49\textwidth]{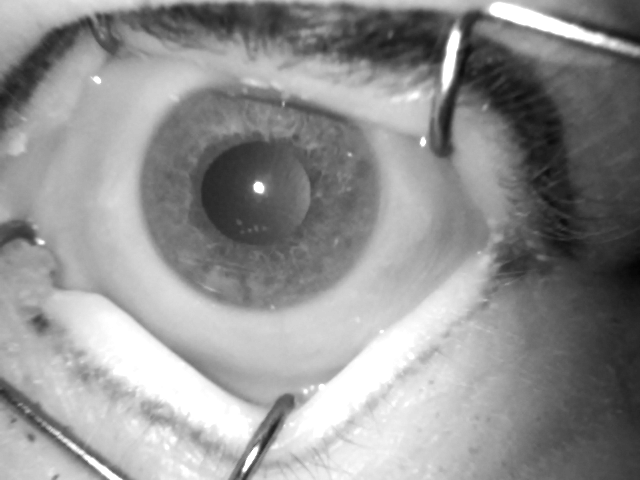}\hskip0.5mm
		\includegraphics[width=0.49\textwidth]{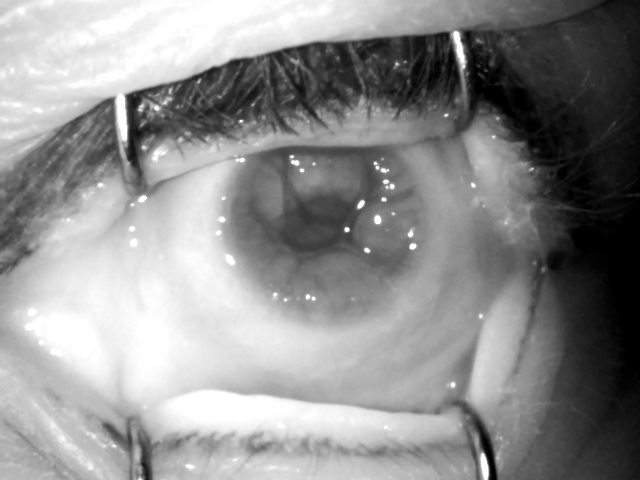}
        \caption{False non-match (1.39)}
    	\end{subfigure}\hskip1.5mm
	\begin{subfigure}[!htb]{0.24\textwidth}
	\centering
		\includegraphics[width=0.49\textwidth]{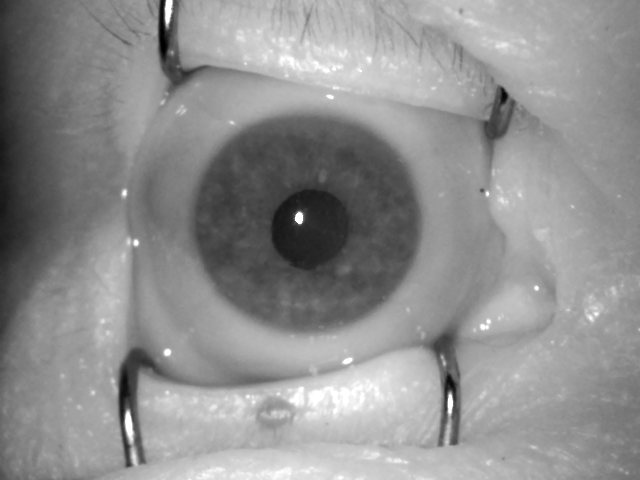}\hskip0.5mm
		\includegraphics[width=0.49\textwidth]{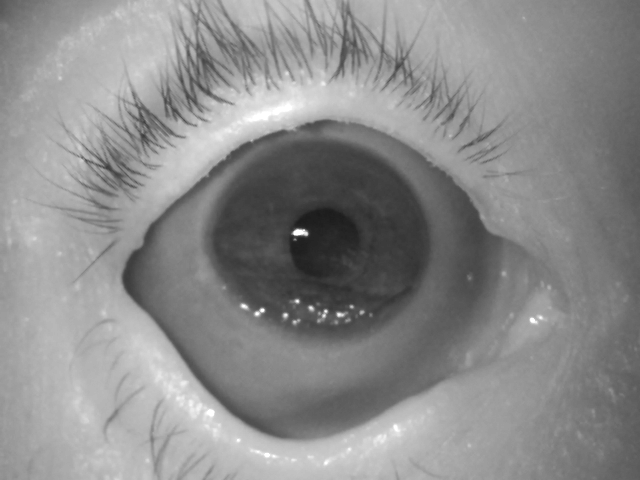}
        \caption{False match (1.09)}
    	\end{subfigure}
	\caption{Same as in Fig. \ref{fig:FM-FNM-OS}, except for the {\bf IriCore} matcher. We assume that IriCore method returns a match when the comparison score is below 1.1.}
	\label{fig:FM-FNM-IC}
\end{figure}

\begin{figure*}[!htb]
	\centering
	\includegraphics[width=0.49\textwidth]{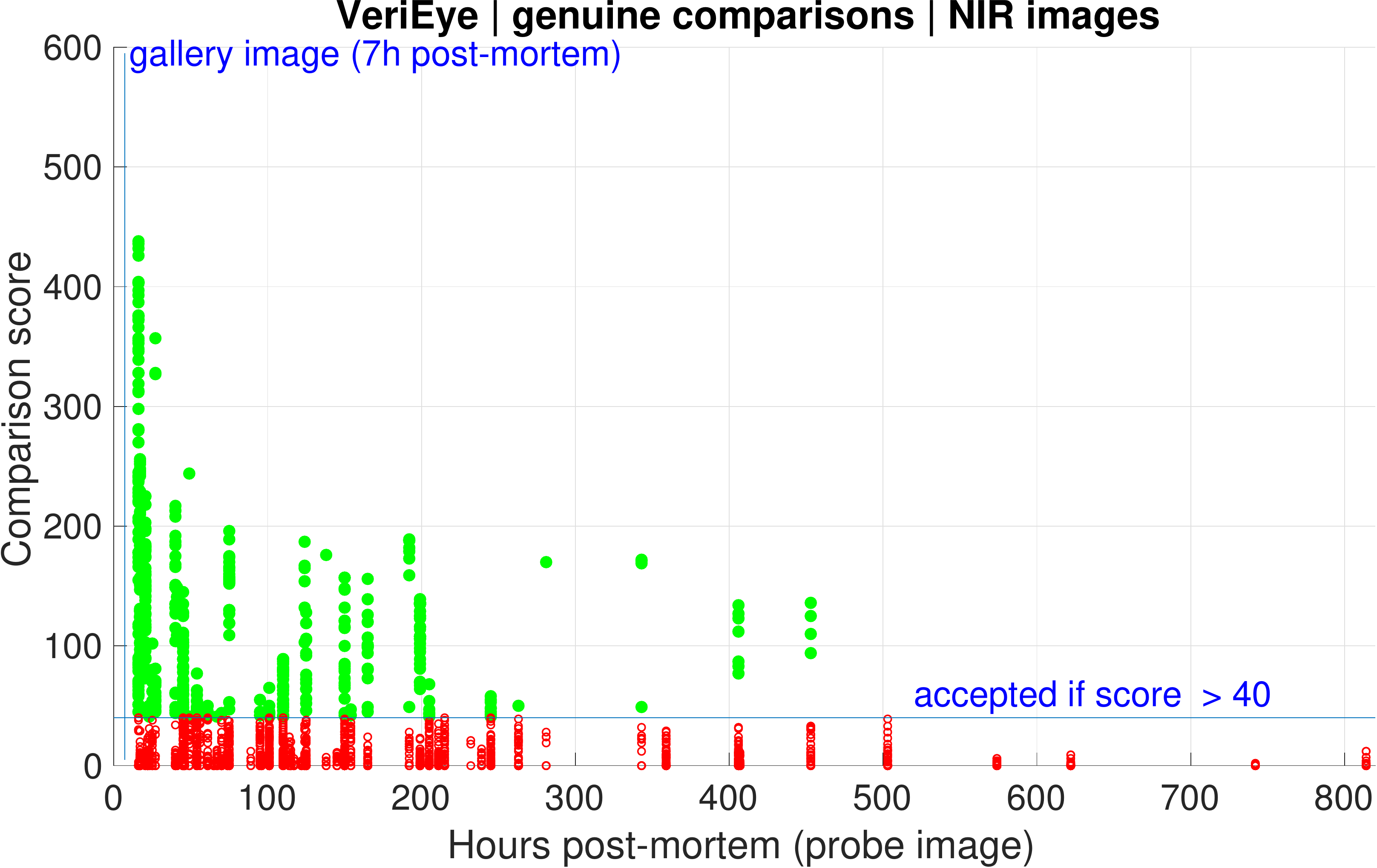}\hfill
	\includegraphics[width=0.49\textwidth]{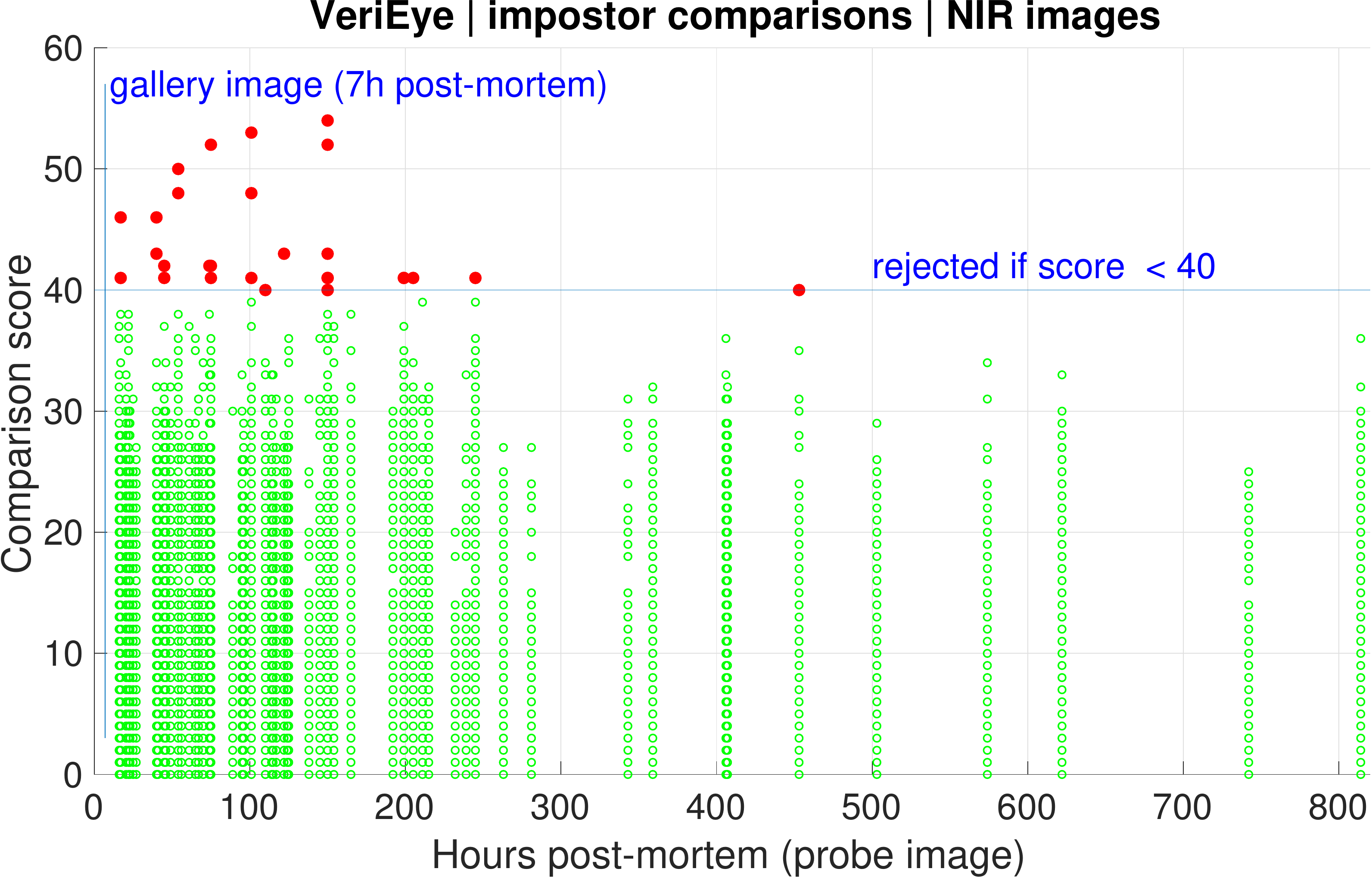}\\\vskip5mm
	\includegraphics[width=0.49\textwidth]{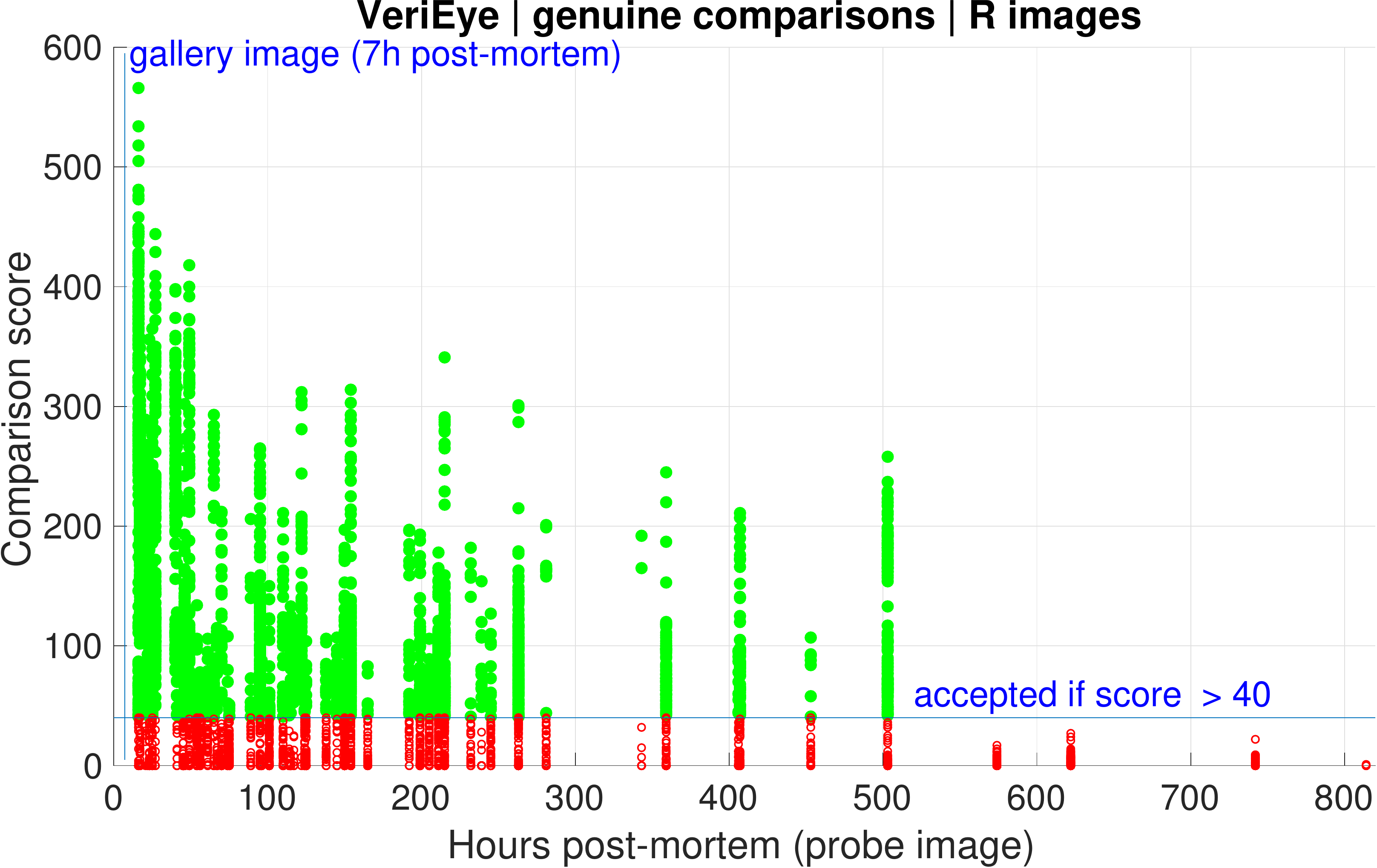}\hfill
	\includegraphics[width=0.49\textwidth]{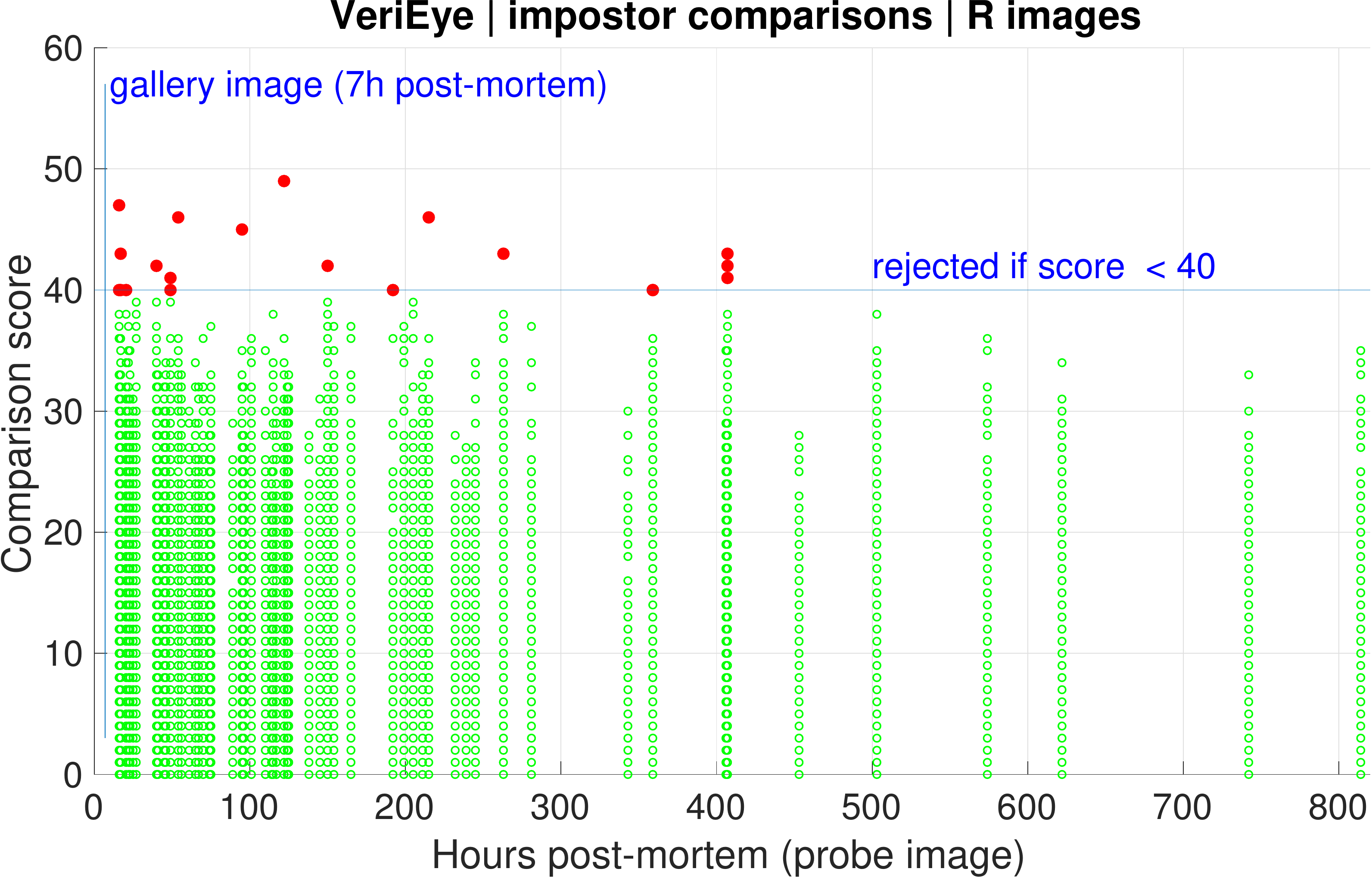}\\	
	\caption{Same as in Fig. \ref{fig:longTerm_OS}, but for the \textbf{VeriEye} matcher.}
	\label{fig:longTerm_VE}
\end{figure*}

\begin{figure}[!htb]
	\centering
	\begin{subfigure}[!htb]{0.24\textwidth}
	\centering
		\includegraphics[width=0.49\textwidth]{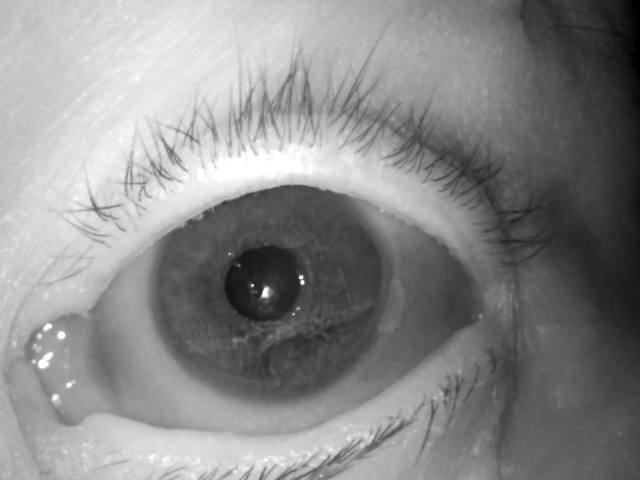}\hskip0.5mm
		\includegraphics[width=0.49\textwidth]{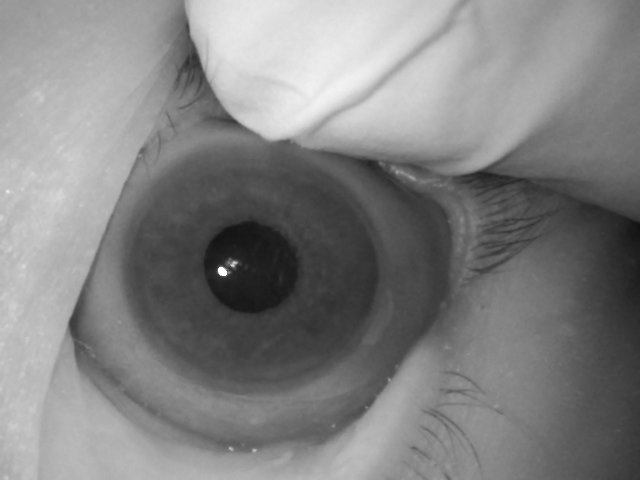}
        \caption{False non-match (0)}
    	\end{subfigure}\hskip1.5mm
	\begin{subfigure}[!htb]{0.24\textwidth}
	\centering
		\includegraphics[width=0.49\textwidth]{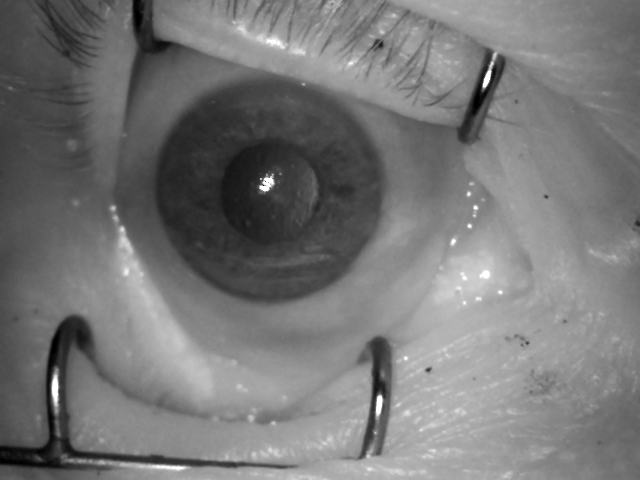}\hskip0.5mm
		\includegraphics[width=0.49\textwidth]{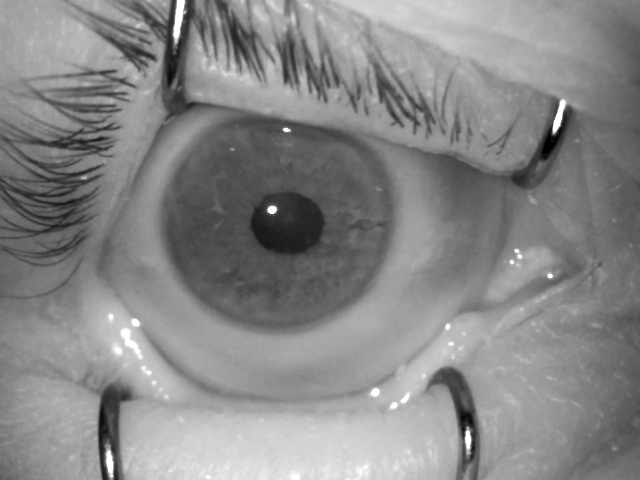}
        \caption{False match (46)}
    	\end{subfigure}
	\caption{Same as in Fig. \ref{fig:FM-FNM-OS}, except for the {\bf VeriEye} matcher. We assume that VeriEye returns a match when the comparison score is above 40.}
	\label{fig:FM-FNM-VE}
\end{figure}

Samples acquired in the second and subsequent sessions are sparsely distributed in time and across the subjects. Thus, Figs. \ref{fig:longTerm_OS} through \ref{fig:FM-FNM-VE} present genuine and impostor scores calculated by all the methods, between session 1 images (5-7 hours after death) and all samples acquired in the following sessions, together with close-up analysis of example false matches and false non-matches. With a difficulty to obtain ante-mortem and then post-mortem iris scans from the same individuals, and after visual inspection of samples (cf. Sec. \ref{sec:VisualInspection}), we assume that iris scans obtained shortly (\ie a few hours) after death would not differ much from those obtained ante-mortem. Thus, images acquired in the first session serve as gallery samples in all subsequent analyses.

In this section we also visually investigate selected image pairs that generated false matches and false non-matches (assuming default acceptance thresholds). Each method, except for OSIRIS with Daugman's score normalization, generated false matches and all matchers generated false non-matches. Thus, we selected the worst pair of images in each case, \ie the most similar images of different eyes and the most distinct images of the same eye, both in terms of the comparison score. We were also able to read the segmentation results in two methods (MIRLIN and OSIRIS), which helped to find a reason behind a given error.

So, how long after death can the iris still offer enough features to generate a correct match? 

For the OSIRIS method, when employing automatic image segmentation, we can expect correct matches for samples obtained up to 150 hours post-mortem for NIR images, and 263 hours for R images, Fig. \ref{fig:longTerm_OS}. However, when the manual corrections to the segmentation stage are introduced, Fig. \ref{fig:longTerm_OSm}, the recognition horizon for NIR samples extends to 263 hours, while for R samples it declines to only 215 hours post-mortem, which may indicate that some of the correct matches obtained for R images were a result of badly segmented irises, and not of a genuine match of iris codes. Plots of impostor comparisons shown in Figs. \ref{fig:longTerm_OS} and \ref{fig:longTerm_OSm} suggest that there is no clear trend in comparison scores when time after death increases, also in the case when manual segmentation was applied.

Figure \ref{fig:FM-FNM-OS} presents the worst results obtained from OSIRIS prior to applying score normalization. Incorrect segmentation explains the reason for a false non-match obtained for the left pair of samples. The right pair and the segmentation results show a very interesting case of a false match. Even with incorrect segmentation, as observed in this case, we still have some `non-occluded' image areas that are significantly different (part of the skin on the first image and part of the wrinkled iris/cornea on the second image). So why the false match is observed? The reason for that may be a very small number of bits being compared due to application of occlusion mask during calculation of the Hamming distance. Note that the occlusions found by the OSIRIS in this image pair are almost mutually exclusive: `non-occluded' part is mostly located on the left part of the hypothetical iris on the first image, while `non-occluded' part of the iris shown on the second image is located mostly on the right. This false match does not happen after applying score normalization that penalizes low numbers of mutually un-occluded bits in OSIRIS codes.

As for the remaining matchers, IriCore and MIRLIN were able to deliver correct matches for samples acquired even 503 hours post-mortem. VeriEye occasionally recognizes samples acquired up to 260 hours after demise, at the assumed acceptance thresholds. IriCore was again a method that rejected a very small number of comparisons (0.11\% of NIR images and 0\% of R images) when compared to other methods (cf. `Long-term analysis' column in Tab. \ref{tab:FTE}). This means that in favorable conditions (IriCore software and IriShield sensor come from the same manufacturer) iris recognition may still be possible {\bf almost 21 days after death}. Unfortunately, as this method does not allow for generating segmentation results, we are not able to investigate whether these correct matches are indeed a result of genuinely matching the corresponding iris features, or of an incorrect segmentation causing the similar portions of the image to be matched.

Figure \ref{fig:FM-FNM-ML} presents iris image pairs yielding false non-match (left pair) and false match (right pair) when the MIRLIN method is used. In this case we can also observe the segmentation results, which give a clear explanation of the observed errors in both cases. A false non-match is caused simply by comparing non-matching iris areas due to incorrect iris localization. A false-match is probably due to comparing sclera that is very similar in both samples, rather than actual iris texture.

Figure \ref{fig:FM-FNM-IC} presents the worst cases leading to false non-match (left pair) and false match (right pair) when the IriCore method is used. The reason for a false non-match is the collapse of the eyeball and severe cornea wrinkling observed in the second sample (acquired 622 hours post-mortem) of the left pair. The right pair of images, however, does not provide any obvious clue for the observed false match.

Figure \ref{fig:FM-FNM-VE} shows the worst pairs of images from the VeriEye method point of view. The possible cause of a false non-match (left pair) is a compensation of lower intraocular pressure by manually pressing the eyeball (a finger of a personnel pressing the eyeball is visible in the second image). This made the cornea less wrinkled (when compared to the first image of the left pair), but simultaneously changed the visible texture, ending up with creation of different iris features for those samples. However, the right pair of images again does not provide a clear explanation for a false-match.

\begin{figure}[!htb]
\centering
\includegraphics[width=0.23\textwidth]{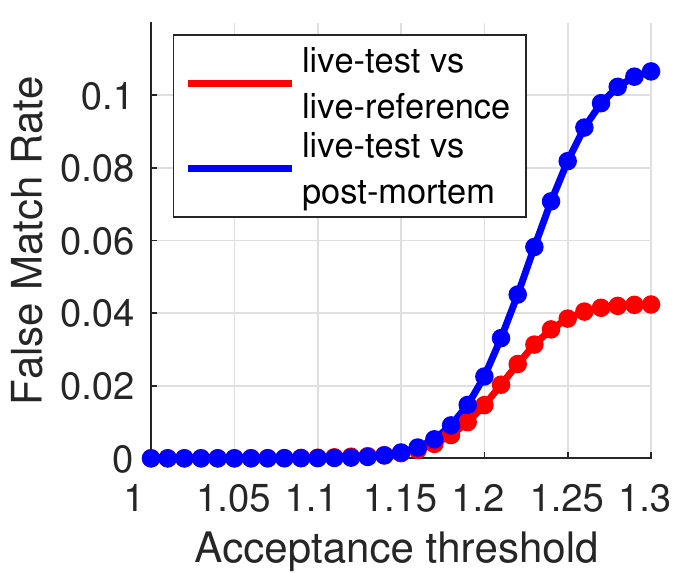}
\includegraphics[width=0.25\textwidth]{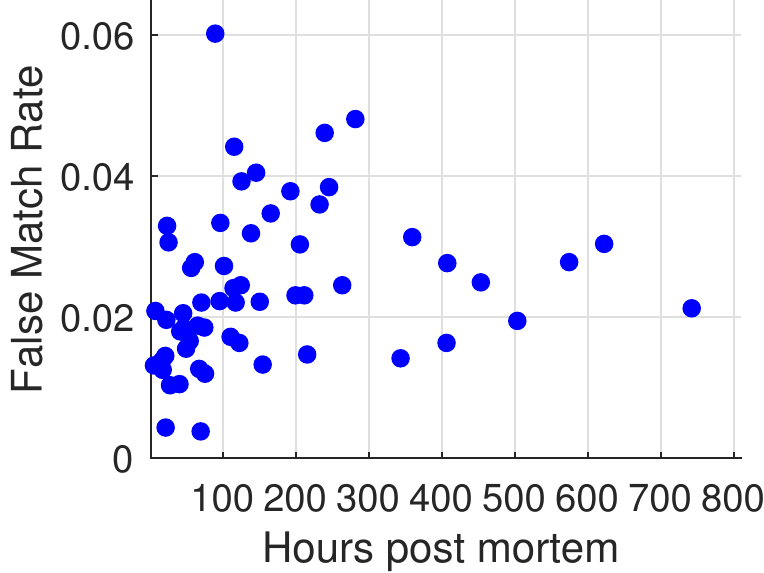}
	\caption{{\bf Left:} FMR as a function of IriCore acceptance threshold for comparisons between live irises and live vs post-mortem irises. {\bf Right:} FMR as a function of post-mortem acquisition time calculated for the IriCore acceptance threshold 1.2, when live irises are compared with post-mortem irises.}
	\label{fig:wwc}
\end{figure}

\subsection{Assessment of False Match Risk When Comparing Post-mortem Samples Against Live-iris Samples}
\label{sec:wwc}

All the above experiments were performed for a closed set of deceased subjects. A question arises: what happens if post-mortem samples are compared in a open-set scenario with live irises of other subjects? To answer this, we used the IriShield MK 2120U sensor again to collect an additional database of iris images from 74 living persons, which was split into subject-disjoint \emph{live-test} and \emph{live-reference} sets, both comprising data from 37 subjects and approximately 600 images. The most accurate (in this study) IriCore method was used to generate and compare two impostor score distributions: (a) scores obtained when matching \emph{post-mortem} iris images (from 37 cadavers, as introduced in this paper) against \emph{live-test} iris images, and (b) scores obtained when matching the same \emph{live-test} images with samples from the subject-disjoint \emph{live-reference} set. The \emph{post-mortem} set was created by randomly choosing images in a way that from each acquisition session half of the samples are selected. This ends up with balanced post-mortem and live iris sets.

Fig. \ref{fig:wwc} (left) presents the False Match Rate calculated for different acceptance thresholds and for two above scenarios (a) and (b). For the threshold equal to 1.1, as recommended by the manufacturer, we observed small FMR $< 0.1$\% in both scenarios. However, when the threshold is relaxed, the chances of getting a false match when live iris images are compared to post-mortem images are higher than in scenario when only live iris images are compared. At the same time, it is hard to see any clear dependency between false match rate and time after death when the iris was photographed, as depicted in Fig. \ref{fig:wwc} (right).

\section{Conclusions}
\label{sec:Conclusions}
This paper delivers the most comprehensive, known to us, analysis of post-mortem iris recognition. The concluding remarks are presented in the next paragraphs in a form of answers to the questions that stimulated this research and were posed in the introductory part. We are also aware of certain limitations of this study, and these are presented at the end of this Section.

\subsection{Question 1: Is automatic iris recognition possible after death?} Post-mortem iris recognition {\bf is possible}. In favorable conditions, when the sensor and the processing methods come from the same manufacturer, we observed EER=2\% when comparing images acquired 5-7 hours post-mortem, collected in visible spectrum with a high-resolution camera, and then converted to grayscale using the R channel. The recognition accuracy can, however, be further improved by employing a human expert to manually annotate the correct iris region in the image, driving the EER down to as low as 1\%, employing the OSIRIS matcher. 

\subsection{Question 2: What are the dynamics of deterioration in iris recognition performance?} Due to inevitable decomposition of human body, recognition accuracy becomes progressively worse. Despite a longer analysis horizon (814 hours) when compared to prior work in this field \cite{BostonPostMortem,TrokielewiczPostMortemICB2016,TrokielewiczPostMortemBTAS2016,PostMortemBoehnenBTAS16}, we did not find any iris image taken later than 503 hours after death that would result in a correct match in all four iris recognition methods used in this work. On the other hand, {\bf 503 hours (almost 21 days)} is a horizon giving an ample amount of time to make post-mortem forensic analysis and suggests that the iris may deliver actual biometric features for a longer period than initially believed. One should also note that four methods used in this study presented highly heterogenous performance on the same set of post-mortem samples. For instance, IriCore failed to compute only 0.11\% comparison scores, while MIRLIN failed to compute 43.51\% comparison scores for NIR images. This may suggest that the algorithms implement significantly different mechanisms to control the quality of iris image pair used to compute the score.

\subsection{Question 3: What type of images are the most favorable for post-mortem iris recognition?} In short-term analysis, when matching samples are obtained shortly after death, we have shown that employing high-resolution R images offers much higher recognition rates when compared to those obtained with NIR images (2\% EER for R images versus 4\% for NIR images, obtained for the IriCore matcher). Both types of images offer close-to-perfect performance when the image segmentation stage is executed manually, with small difference in favor of the R images (1\% ERR versus 2\% for NIR images, obtained for the OSIRIS matcher). This may prove important for forensic applications, which usually involve a human expert, who could then perform the necessary segmentation stage. In addition, we have shown that, probably due to significant differences in the appearance of iris features in post-mortem samples under different wavelengths, a cross-spectral scenario cannot be recommended, with ERRs not dropping below 15\%, even with manual image segmentation in place.

In the long-term analysis, there is no single conclusion on whether NIR or R images are better for post-mortem iris recognition. For the OSIRIS method with manually corrected image segmentation, NIR images seem to offer better chance of getting a correct match. For MIRLIN, R images are generating far more false matches than NIR images, while for IriCore it is the opposite, with NIR images causing more false matches. However, NIR images also allow for correct matches during a longer period post-mortem, compared to R images. VeriEye, on the other hand, seems to work better in general, when R images are used, namely offering less false-matches and more correct matches. Therefore, the matching performance in regard to the image type is heavily matcher-dependent.

\subsection{Question 4: What are the main reasons for errors when comparing post-mortem iris samples?} After visual inspection of the automatic segmentation results for falsely matched and non-matched samples, one of the most obvious reasons for failures is {\bf incorrect localization of iris texture} due to post-mortem decomposition processes. However, most of these failed examples still show the iris texture that is partially or fully useful. It means that new methods in iris segmentation that are insensitive to post-mortem changes might increase the reliability.

\subsection{Question 5: Which factors influence post-mortem iris recognition performance?} The data collected for the purpose of this study represents images acquired from subjects who passed away due to various reasons and of different age. Also, samples for both female and male subjects were collected. This gives a rare opportunity to examine whether cause of death, age and gender can give \emph{a priori} insights on the expected performance of post-mortem iris recognition. Significantly worse comparison scores were observed for subjects who were either poisoned or murdered. Significant differences were observed neither between males and females, nor among groups of comparison scores sorted by the age in the moment of death. However, due to a relatively small number of available genuine comparison scores, these results should be considered as qualitative assessment, rather than formal statistical analysis.

\subsection{Question 6: What are the false-match risks when post-mortem samples are compared against databases of live iris images?}

As shown in Sec. \ref{sec:wwc}, false-match probability may be higher when live iris images are compared with post-mortem samples than when only live samples are used in comparisons. This translates to a higher chance of observing a false match when post-mortem probe sample is compared to a gallery of live iris images, and it calls for post-mortem-specific iris matching strategies to address a possibly of higher false match probabilities in post-mortem iris recognition.

\subsection{Limitations of this research} Although this study brings a few groundbreaking deliveries, and the largest database of post-mortem iris images, one should proceed with the conclusions cautiously and consider this work also as an important call for intensified research in post-mortem iris recognition. The most important limitation that we are aware of is a relatively {\bf small dataset} (37 deceased subjects). This data were collected in a very difficult environment of the hospital mortuary and acquisition moments could not interfere with various examinations, including criminal proceedings. Time spent in the mortuary was beyond our control, hence {\bf irregular acquisition sessions}. The {\bf lack of ante-mortem samples} limits the calculation of the reference templates to the earliest post-mortem images. However, visual inspection suggests that the first-session samples do not exhibit any visible deterioration when compared to living irises. Also, it would be valuable to repeat our analyses for samples collected under varying ambient conditions, as done by Bolme \etal \cite{BolmePostMortemBTAS2016}. However, we are not aware of any database of post-mortem samples different than offered with this paper, and collected under varying environmental conditions, or for more cadaver subjects that would be available to the researchers at the moment of preparation of this paper.

\section*{Acknowledgment}
This study had an institutional review board clearance and the ethical principles of the Helsinki Declaration were carefully followed by the authors. All legal requirements necessary for making the dataset publicly available to researchers have been met. The authors thank Ms Katarzyna Roszczewska and Ms Ewelina Bartuzi for their help with preparation of the ground-truth occlusion masks. The authors also thank Dr. Kevin W. Bowyer who provided valuable comments on an early draft of this paper.


\begin{IEEEbiography}[{\includegraphics[height=1.2in,clip,keepaspectratio]{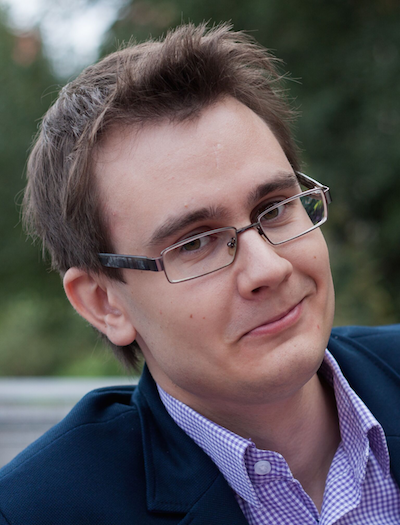}}]{Mateusz Trokielewicz} Received his B.Sc. and M.Sc. in Biomedical Engineering from the Faculty of Mechatronics and the Faculty of Electronics and Information Technology at the Warsaw University of Technology, respectively. He is currently with the Biometrics and Machine Intelligence Lab at the Research and Academic Computer Network and with the Institute of Control and Computation Engineering at the Warsaw University of Technology. His current professional interests include iris biometrics and its reliability against biological processes, iris recognition on mobile devices, and machine learning in biometrics.
\end{IEEEbiography}

\begin{IEEEbiography}[{\includegraphics[height=1.2in,clip,keepaspectratio]{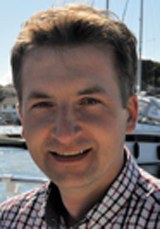}}]{Adam Czajka} PhD, DSc, is an Assistant Professor in the Department of Computer Science and Engineering at the University of Notre Dame. He received M.Sc. in Computer Control Systems and Ph.D. in Biometrics from Warsaw University of Technology (WUT), Poland (both with honors). He is also an Assistant Professor with the Research and Academic Computer Network (NASK), Poland. His scientific interests include biometrics and security, computer vision, and machine learning. Dr Czajka was the Chair of the Biometrics and Machine Learning Laboratory at WUT, the Head of the Postgraduate Studies on Security and Biometrics, and the Vice Chair of the NASK Biometrics Laboratory. He is a Senior Member of the IEEE, an Active Member of the European Association for Biometrics, and an Associate Member of the International Association of Identification.
\end{IEEEbiography}

\begin{IEEEbiography}[{\includegraphics[height=1.2in,clip,keepaspectratio]{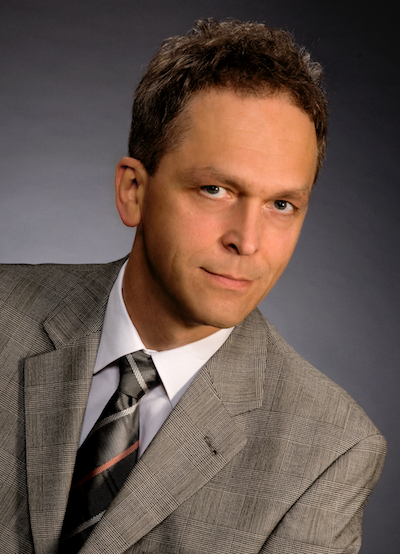}}]{Piotr Maciejewicz} MD, PhD, earned his medical degree at the Medical University of Warsaw, Poland. He completed his residency at the Ophthalmological Clinic in Warsaw, where he is a medical staff member. He is a general ophthalmologist with clinical interests focused on early diagnosis of glaucoma and anterior eye segment problems. Dr. Piotr Maciejewicz is a laser photocoagulation expert in the diabetic retinopathy treatment. He has participated in numerous clinical trials investigating new treatments for retinal diseases such as neovascular age-related macular degeneration involving novel therapeutic agents. He has been elected to prestigious professional organizations including Polish Ophthalmology Society.
\end{IEEEbiography}


\begin{thebibliography}{10}
\providecommand{\url}[1]{#1}
\csname url@samestyle\endcsname
\providecommand{\newblock}{\relax}
\providecommand{\bibinfo}[2]{#2}
\providecommand{\BIBentrySTDinterwordspacing}{\spaceskip=0pt\relax}
\providecommand{\BIBentryALTinterwordstretchfactor}{4}
\providecommand{\BIBentryALTinterwordspacing}{\spaceskip=\fontdimen2\font plus
\BIBentryALTinterwordstretchfactor\fontdimen3\font minus
  \fontdimen4\font\relax}
\providecommand{\BIBforeignlanguage}[2]{{%
\expandafter\ifx\csname l@#1\endcsname\relax
\typeout{** WARNING: IEEEtran.bst: No hyphenation pattern has been}%
\typeout{** loaded for the language `#1'. Using the pattern for}%
\typeout{** the default language instead.}%
\else
\language=\csname l@#1\endcsname
\fi
#2}}
\providecommand{\BIBdecl}{\relax}
\BIBdecl

\bibitem{WaymanJainBiometricSystemsBook2005}
J.~Wayman, A.~Jain, D.~Maltoni, and D.~Maio, ``{An Introduction to Biometric
  Authentication Systems},'' \emph{{Biometric Systems}}, 2005.

\bibitem{JainRossBiometricsForensics2015}
A.~Jain and A.~Ross, ``{Bridging the gap: from biometrics to forensics},''
  \emph{Philisophical Transactions of the Royal Society B}, vol. 370, 2015.

\bibitem{DaugmanPostMortem}
{BBC News}, ``{The eyes have it},'' http://news.bbc.co.uk/2/hi/
  science/nature/1477655.stm, 9 August, 2001 (accessed: April 10, 2016).

\bibitem{SaeedPostMortem}
\BIBentryALTinterwordspacing
A.~Szczepanski, K.~Misztal, and K.~Saeed, ``\BIBforeignlanguage{English}{Pupil
  and iris detection algorithm for near-infrared capture devices},'' in
  \emph{\BIBforeignlanguage{English}{Computer Information Systems and
  Industrial Management}}, ser. Lecture Notes in Computer Science, K.~Saeed and
  V.~Snášel, Eds.\hskip 1em plus 0.5em minus 0.4em\relax Springer Berlin
  Heidelberg, 2014, vol. 8838, pp. 141--150. [Online]. Available:
  \url{http://dx.doi.org/10.1007/978-3-662-45237-0_15}
\BIBentrySTDinterwordspacing

\bibitem{IrisGuardPostMortem}
{IrisGuard}, ``{EyeBank Solution},'' http://www.irisguard.com/
  eyebank/downloads/EyeBankPer\%20Page.pdf (accessed: Jan 2016).

\bibitem{IriTechPostMortem}
{IriTech}, ``{Biometric Access Control — Can Iris Biometric Enhance Better
  Security?}'' http://www.iritech.com/blog/iris-biometric-access-control,
  August 20, 2015 (accessed: January 16, 2016).

\bibitem{BostonPostMortem}
{A. Sansola}, ``{Postmortem iris recognition and its application in human
  identification},'' Master's Thesis, Boston University, 2015.

\bibitem{PostMortemPigs}
S.~K. Saripalle, A.~McLaughlin, R.~Krishna, A.~Ross, and R.~Derakhshani,
  ``{Post-mortem Iris Biometric Analysis in Sus scrofa domesticus},''
  \emph{IEEE 7th International Conference on Biometrics Theory, Applications
  and Systems (BTAS)}, 2015.

\bibitem{RossPostMortem}
A.~Ross, ``{Iris as a Forensic Modality: The Path Forward},''
  http://www.nist.gov/forensics/upload/Ross-Presentation.pdf.

\bibitem{TrokielewiczPostMortemICB2016}
M.~Trokielewicz, A.~Czajka, and P.~Maciejewicz, ``{Post-mortem Human Iris
  Recognition},'' 9th IAPR International Conference on Biometrics (ICB 2016),
  June 13-16, 2016, Halmstad, Sweden, 2016.

\bibitem{TrokielewiczPostMortemBTAS2016}
------, ``{Human Iris Recognition in Post-mortem Subjects: Study and
  Database},'' 8th IEEE International Conference on Biometrics: Theory,
  Applications and Systems, Sep 6-9, 2016, Buffalo, USA, 2016.

\bibitem{WarsawColdIris1}
{Warsaw University of Technology}, ``{Warsaw-BioBase-PostMortem-Iris-v1.0}:
  http://zbum.ia.pw.edu.pl/en/node/46,'' 2016.

\bibitem{BolmePostMortemBTAS2016}
D.~S. Bolme, R.~A. Tokola, and C.~B. Boehnen, ``{Impact of environmental
  factors on biometric matching during human decomposition},'' 8th IEEE
  International Conference on Biometrics: Theory, Applications and Systems
  (BTAS 2016), Sep 6-9, 2016, Buffalo, USA, 2016.

\bibitem{Sauerwein_JFO_2017}
K.~Sauerwein, T.~B. Saul, D.~W. Steadman, and C.~B. Boehnen, ``The effect of
  decomposition on the efficacy of biometrics for positive identification,''
  \emph{Journal of Forensic Sciences}, vol.~62, no.~6, pp. 1599--1602, 2017.

\bibitem{IriCoreIlluminationWavelength}
{U-JIN LED}, ``{ULI-81036A-30IRP IR Lamp Specification,
  http://ujin-led.co.kr/down/ULI-81036A-30IRP.pdf? PHPSESSID =
  2697c4f88777ae81b7ff214684ab0ffd\& ckattempt=1}.''

\bibitem{ISO2}
{ISO/IEC 29794-6}, ``{Information technology -- Biometric sample quality - Part
  6: Iris image data (FDIS)},'' August 2014.

\bibitem{Prieto2015}
\BIBentryALTinterwordspacing
G.~Prieto-Bonete, M.~D. Perez-Carceles, and A.~Luna, ``Morphological and
  histological changes in eye lens: Possible application for estimating
  postmortem interval,'' \emph{Legal Medicine}, vol.~17, no.~6, pp. 437--442,
  2017/08/20 XXXX. [Online]. Available:
  \url{http://dx.doi.org/10.1016/j.legalmed.2015.09.002}
\BIBentrySTDinterwordspacing

\bibitem{LARPKRAJANG20161}
\BIBentryALTinterwordspacing
S.~Larpkrajang, W.~Worasuwannarak, V.~Peonim, J.~Udnoon, and S.~Srisont, ``The
  use of pilocarpine eye drops for estimating the time since death,''
  \emph{Journal of Forensic and Legal Medicine}, vol.~39, pp. 100 -- 103, 2016.
  [Online]. Available:
  \url{http://www.sciencedirect.com/science/article/pii/S1752928X16000093}
\BIBentrySTDinterwordspacing

\bibitem{Belsey2016}
\BIBentryALTinterwordspacing
S.~L. Belsey and R.~J. Flanagan, ``Postmortem biochemistry: Current
  applications,'' \emph{Journal of Forensic and Legal Medicine}, vol.~41, pp.
  49--57, 2017/08/20 XXXX. [Online]. Available:
  \url{http://dx.doi.org/10.1016/j.jflm.2016.04.011}
\BIBentrySTDinterwordspacing

\bibitem{Aslam}
T.~M. Aslam, S.~Z. Tan, and B.~Dhillon, ``Iris recognition in the presence of
  ocular disease,'' \emph{J. R. Soc. Interface 2009}, vol.~6, 2009.

\bibitem{TrokielewiczDiseasesIMAVIS}
M.~Trokielewicz, A.~Czajka, and P.~Maciejewicz, ``{Implications of Ocular
  Pathologies for Iris Recognition Reliability},'' \emph{Image and Vision
  Computing}, 2016.

\bibitem{VeriEye}
Neurotechnology, ``{VeriEye SDK, version 4.3:
  www.neurotechnology.com/verieye.html},'' accessed: August 11, 2015.

\bibitem{ICE2005}
P.~Phillips, K.~Bowyer, P.~Flynn, X.~Liu, and W.~Scruggs, ``{The Iris Challenge
  Evaluation 2005},'' in \emph{2nd IEEE International Conference on Biometrics:
  Theory, Applications and Systems}, 2008.

\bibitem{IREXgeneral}
{National Institute of Standards and Technology (NIST)}, ``{IREX:}
  http://www.nist.gov/itl/iad/ig/irex.cfm,'' accessed on May 20, 2016.

\bibitem{IriCore}
{IriTech Inc.}, ``{IriCore Software Developer’s Manual}, version 3.6,'' 2013.

\bibitem{ISO}
{ISO/IEC 19794-6:2011}, ``{Information technology -- Biometric data interchange
  formats -- Part 6: Iris image data},'' 2011.

\bibitem{MIRLIN}
{Smart Sensors Ltd.}, ``{MIRLIN SDK}, version 2.23,'' 2013.

\bibitem{Monro2007}
D.~M. Monro, S.~Rakshit, and D.~Zhang, ``{DCT-based} iris recognition,''
  \emph{IEEE Transactions on Pattern Analysis and Machine Intelligence --
  Special Issue on Biometrics: Progress and Directions}, vol.~29, no.~4, pp.
  586--595, April 2007.

\bibitem{OSIRIS}
G.~Sutra, B.~Dorizzi, S.~Garcia-Salitcetti, and N.~Othman, ``{A biometric
  reference system for iris. OSIRIS v4.1}:
  http://svnext.it-sudparis.eu/svnview2-eph/ref\textunderscore
  syst/iris\textunderscore osiris\textunderscore v4.1/,'' accessed: Oct 1,
  2014.

\bibitem{Daugman2007NewMethods}
J.~Daugman, ``New methods in iris recognition,'' \emph{IEEE Transactions on
  Systems, Man, and Cybernetics -- Part B: Cybernetics}, vol.~37, no.~5, pp.
  1167--1175, 2007.

\bibitem{TrokielewiczBartuziJTIT}
M.~Trokielewicz and E.~Bartuzi, ``{Cross-spectral Iris Recognition for Mobile
  Applications using High-quality Color Images},'' \emph{Journal of
  Telecommunications and Information Technology}, vol.~3, pp. 91--97, 2016.

\bibitem{TrokielewiczVisibleISBA2016}
M.~Trokielewicz, ``{Iris Recognition with a Database of Iris Images Obtained in
  Visible Light Using Smartphone Camera},'' \emph{2016 IEEE International
  Conference on Identity, Security and Behavior Analysis}.

\bibitem{PostMortemBoehnenBTAS16}
D.~S. Bolme, R.~A. Tokola, C.~B. Boehnen, T.~B. Saul, K.~A. Sauerwein, and
  D.~W. Steadman, ``Impact of environmental factors on biometric matching
  during human decomposition,'' pp. 1--8, 2016.

\end{thebibliography}
\end{document}